\documentclass[journal,twoside,web]{ieeecolor}
\usepackage{generic}
\usepackage{cite}
\usepackage{amsmath,amssymb,amsfonts}
\usepackage{graphicx}
\usepackage{algorithm,algpseudocode}
\usepackage{hyperref}
\hypersetup{hidelinks=true}
\usepackage{textcomp}
\usepackage{array}
\usepackage{soul}
\usepackage{booktabs}
\usepackage{multirow}
\usepackage{tablefootnote}
\usepackage{subfigure}
\usepackage{bbm, dsfont}
\usepackage{stfloats}

\newcommand{\figref}[1]{Fig.~\ref{fig:#1}}

\usepackage{colortbl}%
  \newcommand{\myrowcolour}{\rowcolor[gray]{0.925}}

\newcolumntype{C}[1]{>{\centering\arraybackslash}m{#1}}

\usepackage[T1]{fontenc}
\usepackage{tabularx}
\usepackage{amsopn}
\usepackage{listings}
\usepackage{comment}
\usepackage{commath}
\usepackage[mathcal]{euscript}
\usepackage[table]{xcolor}
\usepackage{nicematrix}
\usepackage{fancybox}

\def\BibTeX{{\rm B\kern-.05em{\sc i\kern-.025em b}\kern-.08em
    T\kern-.1667em\lower.7ex\hbox{E}\kern-.125emX}}
{Arefeen \MakeLowercase{\textit{et al.}}: GlyTwin: Enhancing Digital Twin for Glucose Control in Type 1 Diabetes using Patient-Centric Counterfactual Treatments}
\begin{document}
\title{GlyTwin: Enhancing Digital Twin for Glucose Control in Type 1 Diabetes using Patient-Centric Counterfactual Treatments}
\author{Asiful Arefeen\textsuperscript{1,2}*, Saman Khamesian\textsuperscript{1,2}, Maria Adela Grando\textsuperscript{1}, Bithika Thompson\textsuperscript{3}, Hassan Ghasemzadeh\textsuperscript{1}
\thanks{\textsuperscript{1}College of Health Solutions, Arizona State University, Phoenix, AZ 85004, USA}
\thanks{\textsuperscript{2}School of Computing and Augmented Intelligence, Arizona State University, Tempe, AZ 85281, USA}
\thanks{\textsuperscript{3}Department of Endocrinology, Mayo Clinic Arizona, Scottsdale, AZ 85259, USA}
\thanks{*\textbf{Corresponding author}: aarefeen@asu.edu}
\thanks{**\textbf{Code available at:} \href{https://github.com/Arefeen06088/GlyTwin}{\textcolor{blue}{github.com/Arefeen06088/GlyTwin}}}
}

\maketitle

\begin{abstract}
Frequent and long-term exposure to hyperglycemia increases the risk of chronic complications, neuropathy, nephropathy, and cardiovascular disease. Existing continuous subcutaneous insulin infusion (CSII) and continuous glucose monitoring (CGM) technologies can only model specific aspects of glycemic regulation—like predicting hypoglycemia and administering small insulin boluses. Similarly, current digital twin approaches in diabetes management are primarily focused on predicting glucose response to human behavior and insulin therapy. As a result, current technologies lack the ability to provide alternative treatment scenarios that could guide proactive behavioral interventions for optimal diabetes management. To address this gap, we propose \textbf{\textit{GlyTwin}}\textsuperscript{**}, a novel computational framework that enhances capabilities of digital twin technologies by integrating counterfactual explanations to simulate optimal behavioral treatments for glucose control. GlyTwin generates counterfactual treatments by recommending adjustments to behavioral choices such as carbohydrate intake and insulin dosing to significantly reduce the occurrences and duration of hyperglycemic events. Additionally, GlyTwin incorporates stakeholders' preferences into its intervention-generation process and ensures that the tool itself is personalized and user-centric. We evaluate GlyTwin on AZT1D, a new dataset that we have constructed by collecting longitudinal data from $50$ individuals living with type 1 diabetes (T1D) on automated insulin delivery (AID) systems, each monitored for $26$ days. Results show that GlyTwin outperforms state-of-the-art methods for generating counterfactual explanations with $85.8\%$ valid explanations and $87.3\%$ effectiveness in preventing hyperglycemia when compared against historical data.
\end{abstract}


\section{Introduction}
\label{sec:introduction}
\IEEEPARstart{T}{ype} 1 diabetes (T1D) has a significant economic burden. In 2018, a person with T1D spent \$25,652 annually on diabetes management \cite{Reynolds2023CostAU} in the United States. As the body does not produce any insulin, individuals with T1D require insulin treatment to survive. However, insulin dosing is complicated and requires constant decision-making by the end-user or technology regarding the amount of meal intake \cite{Arefeen2025MealMeterUM}, as well as the appropriate timing and dosage of insulin administration. As a result, individuals with T1D often experience abnormal glucose events, such as hypoglycemia and hyperglycemia, which occur when blood glucose levels fall below 70 mg/dL or rise above 180 mg/dL, respectively \cite{ElSayed20226GT}. Individuals with T1D often face challenges in glucose control that complicate the disease over time. Maintaining within-target glycemic range without significant hyperglycemia and hypoglycemia is challenging. Even with the advent of continuous glucose monitor (CGM) and automated insulin delivery (AID) systems, only 64.1\% of individuals with T1D using both technologies are able to achieve the recommended glycemic targets \cite{Sherr2024SevereHA}. Nonetheless, AI-driven interventions targeting dysglycemia have the potential to improve HbA1c, insulin resistance, fasting glucose, glycemic control, and promote weight loss, as well as reduce the disease burden in individuals with T1D \cite{Shamanna2024OneyearOO, Lee2023AnID}.

An emerging but underutilized technology in this context is digital twin, a growing technological paradigm that can model physiological processes and simulate treatments to support clinical decision-making. The current utility of digital twin technologies in behavioral health remains largely confined to predictive modeling of future health outcomes \cite{Cappon2023ReplayBGAD, Cappon2023SystemAO, Surian2024ADT, Silfvergren2021DigitalTP} and lacks mechanisms to actively guide personalized behavioral interventions. To maximize the potential of digital twin technologies in behavioral health, AI-driven intervention systems must transcend mere modeling and factor analysis to deliver precise, actionable recommendations tailored to the user’s context to prevent adverse health outcomes and improve health. For individuals with T1D, this may include specific guidance on step counts, exercise duration and intensity, nutrient intake, macronutrient composition of food, and insulin timing and dosage to avoid dysglycemia. To the best of our knowledge, a noticeable gap persists in digital twin research to identify optimal behavioral pathways that can prevent adverse glucose outcomes. The proposed work integrates predictive modeling with novel methods of generating counterfactual explanations (CFs) to provide real-time treatments for dysglycemia prevention and diabetes management.

\begin{figure}[!h]
\centerline{\includegraphics[width=\columnwidth]{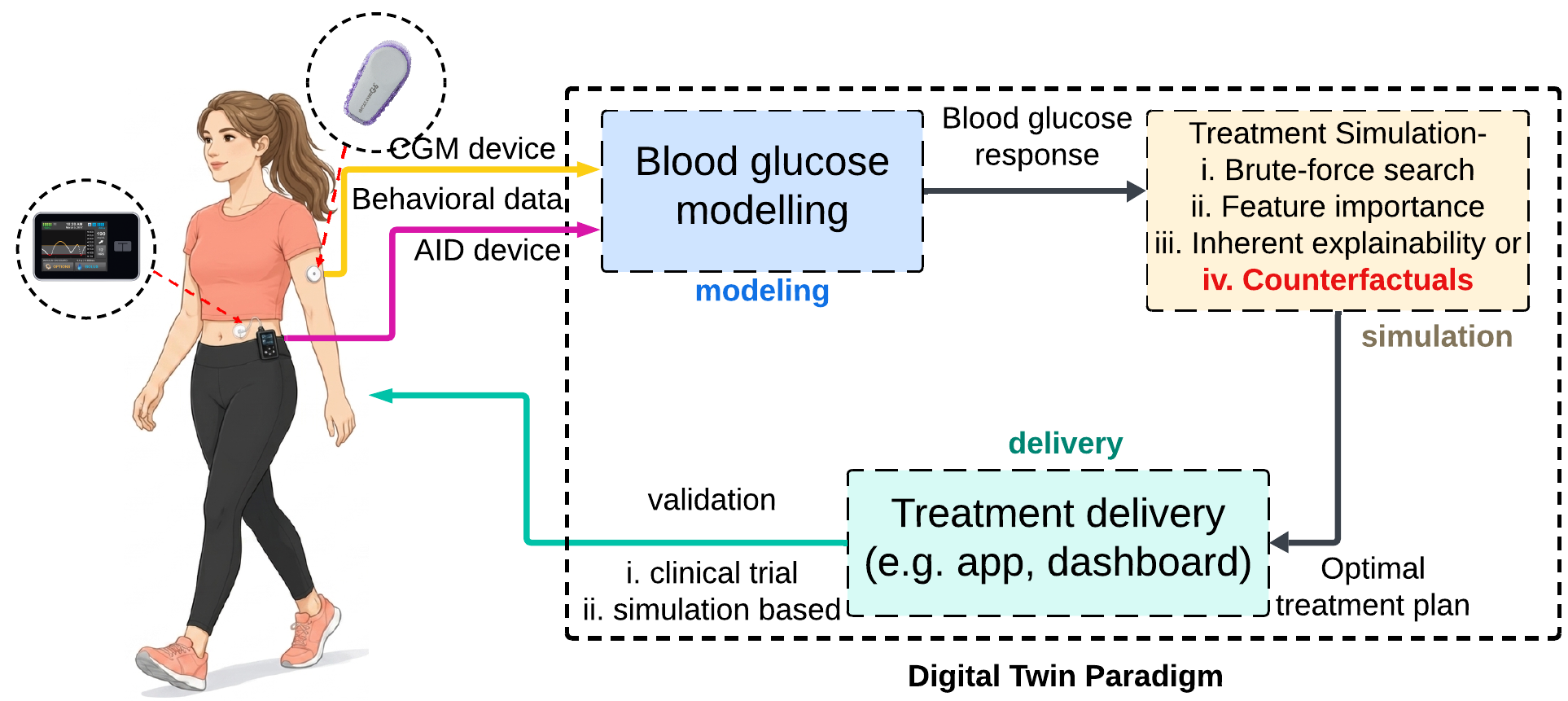}}
\caption{Digital twin framework with enhanced capabilities that can model physiological response, simulate treatments, and identify optimal treatment.}
\label{fig:digitwin}
\end{figure}

Our proposition, \textbf{\textit{GlyTwin}}, is built upon this vision. Following \figref{digitwin}, we envision that a digital twin framework for glucose control in T1D comprises three main pillars focused on modeling of blood glucose response, simulation of behavioral treatments, and delivery of optimal treatments. Prior research has primarily focused on blood glucose modeling by designing machine learning models that predict blood glucose levels based on past behavioral and physiological data \cite{Cappon2023ReplayBGAD, Cappon2023SystemAO, Surian2024ADT, Arefeen2023GlySimMA, Shroff2023GlucoseAssistPB, Khamesian2025GlycemicAwareAA, Farahmand2025AttenGlucoMT, Machiraju2025TimeAwareCF, Soumma2026GlyRAGCR}. While a machine learning model can be potentially used to generate simulated interventions following a brute-force search by modifying the inputs of the model and observing the glucose response at the model output, the number of candidate treatments under this approach is exponentially high. Specifically, for a machine learning model with $n$ inputs, there are $2^n$ permutations of the inputs that one can modify in order to generate an output for the model. Additionally, for each permutation of the inputs, there could be an exponential number of different values that each input permutation can take. Although clinical decision-making rarely requires such exhaustive search, clinicians typically compare only a small set of plausible adjustments. In contrast, our approach for identifying optimal treatments relies on generating CFs, a computationally-efficient machine learning approach to investigate how a desired outcome from a model (e.g., normal glucose range) can be obtained by generating new feature inputs (e.g., carbohydrate intake, exercise, insulin time and amount).

We propose a mechanism to generate CF reasons, use them as a means to provide personalized behavioral recommendations beyond just predictive modeling \cite{Faruqui2024NurseintheLoopAI}, and integrate intervention planning to digital twin systems by identifying minimal behavioral modifications that lead to in-range postprandial blood glucose response through CFs. Therefore, as shown in \figref{digitwin}, the enhanced paradigm can model physiological outcomes, simulate and identify optimal behavioral treatment pathways. This simulation can be done via optimal outcome search, using inherently explainable models (e.g. decision tree, logistic regression) or post-hoc explainable AI (XAI) methods. However, outcome search in high-dimensional spaces is challenging, and interpretable models may yield incomplete guidance \cite{Arefeen2022ForewarningPH}, while feature importance based methods may lack the granularity desired in intervention design. For example, traditional XAI is more interested in identifying features most influential behind a prediction. As a result, the ability of XAI techniques like LIME \cite{Ribeiro2016WhySI}, TIME \cite{Sood2021FeatureIE}, SHAP \cite{Lundberg2017AUA} and others \cite{Zhou2019UnbiasedMO,Fisher2018AllMA,Bento2020TimeSHAPER,Tonekaboni2020WhatWW,A2023ASR} are limited to creating a hierarchy of the most relevant input features from the model's perspective. Often, these explanations are provided in view of low-level features that are hardly understandable to end-users \cite{Barbiero2021EntropybasedLE}, which undermines the main objective of XAI. Feature relevance fosters trust in a model \cite{Caruana2015IntelligibleMF}. However, when it comes to implementing treatment plan in digital health, designing interventions requires more granular explanations.

\begin{figure}[!h]
\centerline{\includegraphics[width=0.68\columnwidth]{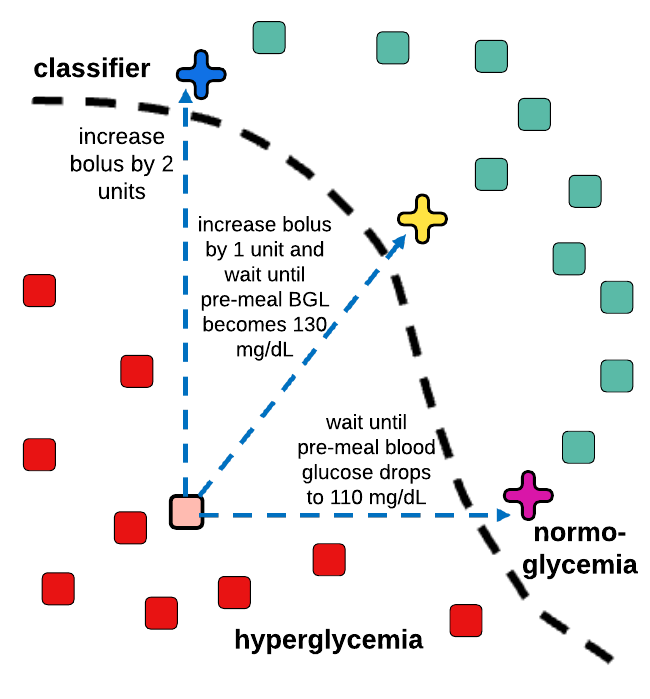}}
\caption{Illustration on how counterfactual reasoning can suggest multiple behavioral adjustments, like modifying insulin dose or meal timing or both, to transform a hyperglycemic event into a normoglycemic outcome.}
\label{fig:fig1}
\end{figure}

To provide granularity in interventions, CFs can serve as a feasible choice for treatment simulation. CF is a more targeted branch of XAI that instills trust in model by describing the smallest change to the feature values that alters the prediction to a desired output. Alternatively, the explanations themselves can be used as interventions to prevent adverse events. \figref{fig1} illustrates how CF reasoning is used to prevent hyperglycemia, showing a decision boundary that separates hyperglycemic (red squares) and normoglycemic (green squares) outcomes. For an observed hyperglycemic event (pink square), GlyTwin suggests actionable behavioral CFs: (i) increasing the pre-meal insulin bolus by $2$ units (blue cross), (ii) increasing the bolus by $1$ unit combined with waiting until the pre-meal BGL decreases to $130$ mg/dL (yellow cross), and (iii) waiting until the pre-meal BGL drops to $110$ mg/dL without modifying the bolus (purple cross). The blue arrows demonstrate the behavioral adjustment pathways to transition from the observed hyperglycemic event to each of these CF normoglycemic scenarios.

Importantly, although \figref{fig1} illustrates multiple behavioral pathways for preventing postprandial hyperglycemia, all generated trajectories reflect clinically reasonable and physiologically coherent actions consistent with typical endocrinology practice. The interventions are aligned with current clinical guidelines—based on clinician feedback incorporated into the design of GlyTwin—and are therefore intended to support safety, feasibility from a user’s standpoint, and practical actionability. The optimization safeguards ensure that the intervention pathways are not only algorithmically optimal but also medically appropriate for T1D management. Thus, GlyTwin accommodates user preferences and behavioral flexibility while offering safe and actionable pathways to avoid hyperglycemia.

CFs can be either actual instances from the training data \cite{Brughmans2021NICEAA, Mamun2024UseOW}, or hypothetical synthetic samples made with a combination of feature values \cite{Arefeen2023DesigningUB, Arefeen2024GlyManGM}. The utility of CF interventions in diabetes research is not entirely new. Lenatti et al. showed that CFs significantly improve fasting blood sugar, systolic blood pressure, triglycerides and HDL among people at risk for diabetes \cite{Lenatti2022ANM}. In a separate study, authors generated CF recommendations related to a healthy lifestyle for preventing diabetes onset \cite{Lenatti2022CharacterizationOT}. Xiang et al. produced realistic CFs for diabetes prevention leveraging variational autoencoders \cite{Xiang2022RealisticCE}. Shah et al. \cite{Shah2025EnhancingMS} used CFs to generate alternate cases where metabolic syndrome is absent. Soumma et al. \cite{Soumma2026CounterfactualMW, Soumma2025SenseCFLC} generated LLM-based CFs for data augmentation. However, no prior research integrated user preferences into CF generation for glucose control. Hence, the actionable insights are often infeasible, unrealistic and contradictory to domain knowledge \cite{Bhattacharya2023DirectiveEF}.

GlyTwin aims to enhance the scope of digital twin in preventing adverse health outcomes with CF interventions as a novel addition. While existing digital twin paradigm is primarily focused on projecting glycemic responses to assist with intervention planning, GlyTwin explores behavioral trajectories and creates intervention plans based on modifiable factors. Additionally, GlyTwin takes stakeholders' preferences - like those of individuals with T1D and physicians - by weighting features in the CF generation process to reflect individual priorities. Thereby, GlyTwin ensures tailored intervention plans that enhance within-target glycemic range and user trust. Contributions made through the design of GlyTwin can be shortlisted as follows.

\begin{itemize}
    \item GlyTwin equips digital twin systems with CFs, thus providing a means for treatment simulations. Using a novel model-agnostic and user-centric algorithm, GlyTwin generates interventions aimed at preventing postprandial hyperglycemia through behavioral modifications.

    \item The interventions provided by GlyTwin are personalized, meaning that they reflect individuals' preferences to withhold certain feature changes and operate within the individuals' limitations in terms of behavioral preferences.

    \item GlyTwin is developed and tested using a new clinical dataset collected in free-living conditions and a competitive analysis is drawn against existing methods using standard validation metrics.
\end{itemize}

\section{Problem statement}

Assume that $\mathcal{D}$ = \{($X_1$,$y_1$), ($X_2$, $y_2$), $\dots$ , ($X_n$, $y_n$)\} be a dataset of $n$ instances that has longitudinal health observations related to eating events and the corresponding health outcome such as blood glucose level categories. Each instance $X_i$ = [${x_i}^{1}$, ${x_i}^{2}$, $\dots$, ${x_i}^{d}$] consists of $d$ features including actionable behavioral features (e.g., diet, medication) and non-actionable features (e.g., age, gender, A1C). Considering $c$ possible classes for health outcome $Y$, where $y_i \in [1,c]$, a probabilistic AI model or classifier $f$ can be trained to map the $d$-dimensional input features to the $c$ classes and give us their corresponding prediction probabilities $f_1, f_2, \dots,f_c$:
\begin{equation*}
f : \mathbb{R}^{d} \rightarrow [1,c]
\end{equation*}

Given a test sample $X_T$ predicted to indicate post-prandial hyperglycemia (i.g., ${\mbox{argmax}}f(X_T)$ = $hyperglycemia$), a key question emerges: how to develop an effective intervention plan that empowers the end-user to make informed behavioral changes to prevent the impending hyperglycemia while also preserving their preferences simultaneously?

\section{Problem solution}

To generate CFs, we have to go through several constraints and satisfy them. For example, the CFs must belong to the desired class, must not change too much from the factuals and must reflect user preferences. We assume that the stakeholder's preferences for behavior changes are represented in vector $R(X_T)$ = \{$r_1$, $\dots$, $r_d$\}, where each $r_i \in [0,1]$ represents the relative preference of the $i$-th feature for modification during intervention. Specifically, a value of $r_i = 1$ indicates that the stakeholder is strongly in favor of modifying the $i$-th feature, while $r_i = 0$ implies no preference for modification. Our goal in GlyTwin is to generate CFs $(X_T^*)$ that satisfy the following criteria:
\begin{itemize}
    \item \textit{\textbf{interventional}}: $X_T^*$ must change the class of the initial prediction from hyperglycemia to normoglycemia;
    \item \textit{\textbf{minimal}}: $X_T^*$ must be minimally distant from the hyperglycemic factual sample $X_T$;
    \item \textit{\textbf{partial}}: $X_T^*$ must favor stakeholders' preferences expressed in feature weights $R$ and
    \item \textit{\textbf{plausible}}: $X_T^*$ must be realistic, i.e., the features of the CFs must fall within the distribution of the dataset $\mathcal{D}$.
\end{itemize}
We formulate the CF generation process using a multi-objective optimization problem as shown in Equation \eqref{optimization}, where the interventional, minimal, partial, and plausible requirements are formalized in the first to third terms, respectively.

\begin{equation}
\begin{aligned}
\label{optimization}
\min_{X_T^*} \Big[ &CE\big(f_n(X_T^*), \overrightarrow{n}\big) + (1-R) \odot |X_T^* - X_T| + d\big(X_T^*, X\big) \Big] \\
\end{aligned}
\end{equation}

Here, $CE(\cdot)$ is the cross-entropy loss between model's prediction on the CF and normoglycemia, $d(\cdot)$ is the distance function.

For any test sample $X_T \in \mathbb{R}^{d}$ classified as \textit{hyperglycemia}, the key idea would be finding the smallest adversarial perturbation $\delta_{min,p}$, that can be added to $X_T$, such that the perturbed point $X_T +\delta$ remains within a specified set of constraints $C$ and the classifier decision changes to $normoglycemia$. Therefore, the minimal adversarial perturbation for $X_T$ with respect to the $l_p$-norm can be defined mathematically-
\begin{align*}
    \delta_{min,p} &= \min_{\delta \in \mathbb{R}^d}||\delta||_p\\
    \text{s. t.  }&f_n(X_T +\delta)>\max_{h\neq n, \forall a} f_h(X_T +\delta),\text{ } X_T +\delta\in C
\end{align*}

To solve the optimization problem (\ref{optimization}) using perturbation, we employ an iterative approach, where we adjust the features of $X_T$ step-by-step based on the saliency scores and stakeholder preferences, while keeping the changes within realistic bounds.

To ensure minimal changes to the factual samples, feature saliency is calculated for each modifiable feature. While the magnitude of feature saliency represents the impact of changing a specific feature on the model's prediction for the target class, its polarity reveals the direction of perturbation. Therefore, identifying the most salient feature on each iteration helps GlyTwin determine which feature, when modified, will provide strongest effect towards the desired outcome.

For each modifiable feature in $x_{mod}$, the saliency score $S(x_T,y',i)$ is calculated by perturbing the feature value by a small amount and observing the change in the model's prediction probability for the target class. This change is captured through the forward derivative of the prediction with respect to the feature,
\begin{equation}
    S(x_T, y', i) = \frac{f_{n}(x_T^{*i} + \delta_i) - f_{n}(x_T^{*i})}{\delta_i} \quad \forall x_T^{*i} \in x_{mod}
\end{equation}

Leveraging the feature saliency, along with stakeholders' preference weights, a combined score is calculated to determine the feature to be changed. Specifically, the combined score $C_i$ for each feature $x_T^i$ is computed by adding the normalized saliency score $S(x_T^*, y', i)'$ (normalized to the range $\left[-1, 1\right]$) to the sum of the physician's and the user's preference weights ($w_p$ and $w_u$), which later helps determine the feature to modify-
\begin{align*}
i^\prime = \arg\max_i \bigl[\left|S(x_T^*, y', i)'\right| + (w_p + w_u)\bigr]
\end{align*}
\begin{align*}
x_T^{*i^\prime} = x_T^{*i^\prime} + \delta_i \cdot \text{sign}&\left[S(x_T^*, y', i)'\right] \\
\text{subject to,} \quad \mathfrak{f}_{\min}[i^\prime] \leq &x_T^{*i^\prime} \leq \mathfrak{f}_{\max}[i^\prime]
\end{align*}

$i^\prime$ denotes the index of the feature with the highest combined score. With this approach the feature selected for modification is both highly salient for the model and aligned with the stakeholders' preferences. Next, we increment the selected feature with preset step size $\delta_i$ towards the direction given by saliency. While doing so, we make sure the feature value is bounded below and above by predetermined limits, $\mathfrak{f}_{min}[i^\prime]$ and $\mathfrak{f}_{max}[i^\prime]$, respectively.

Finally, we add a \textit{stopping criteria} to ensure that the algorithm terminates when it continues to make no improvements on the target class prediction for several rounds.

Algorithm~\ref{alg} executes this iteration-based intervention search, ensuring that minimum and within-range changes are made to features that have the highest combined score of feature saliency and stakeholder weights. We propose two variants of GlyTwin: one that allows adjusting the time between bolus and meal intake in either direction (\textit{\textbf{GlyTwin bi-$\Delta t$}}), and another that restricts adjustments to a single direction and ensures the bolus is suggested only before the meal (\textit{\textbf{GlyTwin one-$\Delta t$})}.

\begin{algorithm}[!h]
\caption{Generating Counterfactual explanations with GlyTwin}\label{alg}
\begin{algorithmic}[1]
\Require Original observation $X_T=[x_T^1, \dots, x_T^d]$, target class $y'$, model $f$, maximum iterations $N$, target confidence for normoglycemia $\gamma$, perturbation/step sizes $\delta=[\delta_1,\dots,\delta_d]$, indices of modifiable features $I_{mod}$, feature min values $\mathfrak{f}_{min}$, feature max values $\mathfrak{f}_{max}$, physician's preference weights $w_p$, user's preference weights $w_u$, multiplier $\mathcal{M}=[1,\dots,1]_{d}$
\Ensure Counterfactual $X_T^* = [x_T^{*1}, \dots, x_T^{*d}]$
\State $X_T^* \gets X_T$
\State $n \gets 0$ \Comment{Track the number of iterations}

\While{$f(X_T^*)[y'] < \gamma$ \textbf{and} $n < N$}
    \State $S \gets [0,\dots,0]_{d}$ \Comment{Initialize saliency scores}
    \State $C \gets [0,\dots,0]_{d}$ \Comment{Combined score for features}
    \For{ $i \in I_{mod}$}
        \State $X_T^{*'} \gets X_T*$
        \State $x_T^{*i'} \gets x_T^{*i'} + \delta_i$
        \State $S_i \gets \frac{f_n(X_T^{*'})[y'] - f_n(X_T^*)[y']}{\delta_i}$ \Comment{Feature saliency}
        \State $C_i \gets \left[|S_i| + w_p[i] + w_u[i]\right]\cdot\mathcal{M}[i]$
    \EndFor
    \State $i \gets \arg\max(C) $

    \State $x_T^{*i} \gets x_T^{*i} + \text{sign}(S_i) \cdot \delta_i$
    \If{$x_T^{*i} < \mathfrak{f}_{min}[i]$}
        \State $x_T^{*i} \gets \mathfrak{f}_{min}[i]$
        \State $\mathcal{M}[i] \gets 0$
    \ElsIf{$x_T^{*i} > \mathfrak{f}_{max}[i]$}
        \State $x_T^{*i} \gets \mathfrak{f}_{max}[i]$
        \State $\mathcal{M}[i] \gets 0$
    \EndIf
    \State $n \gets n + 1$
    \If{stopping criteria is met}
    \State \textbf{break}
    \EndIf
\EndWhile
\State \Return $X_T^*$
\end{algorithmic}
\end{algorithm}

\begin{figure*}[!t]
\centering
    \includegraphics[width=0.9\linewidth]{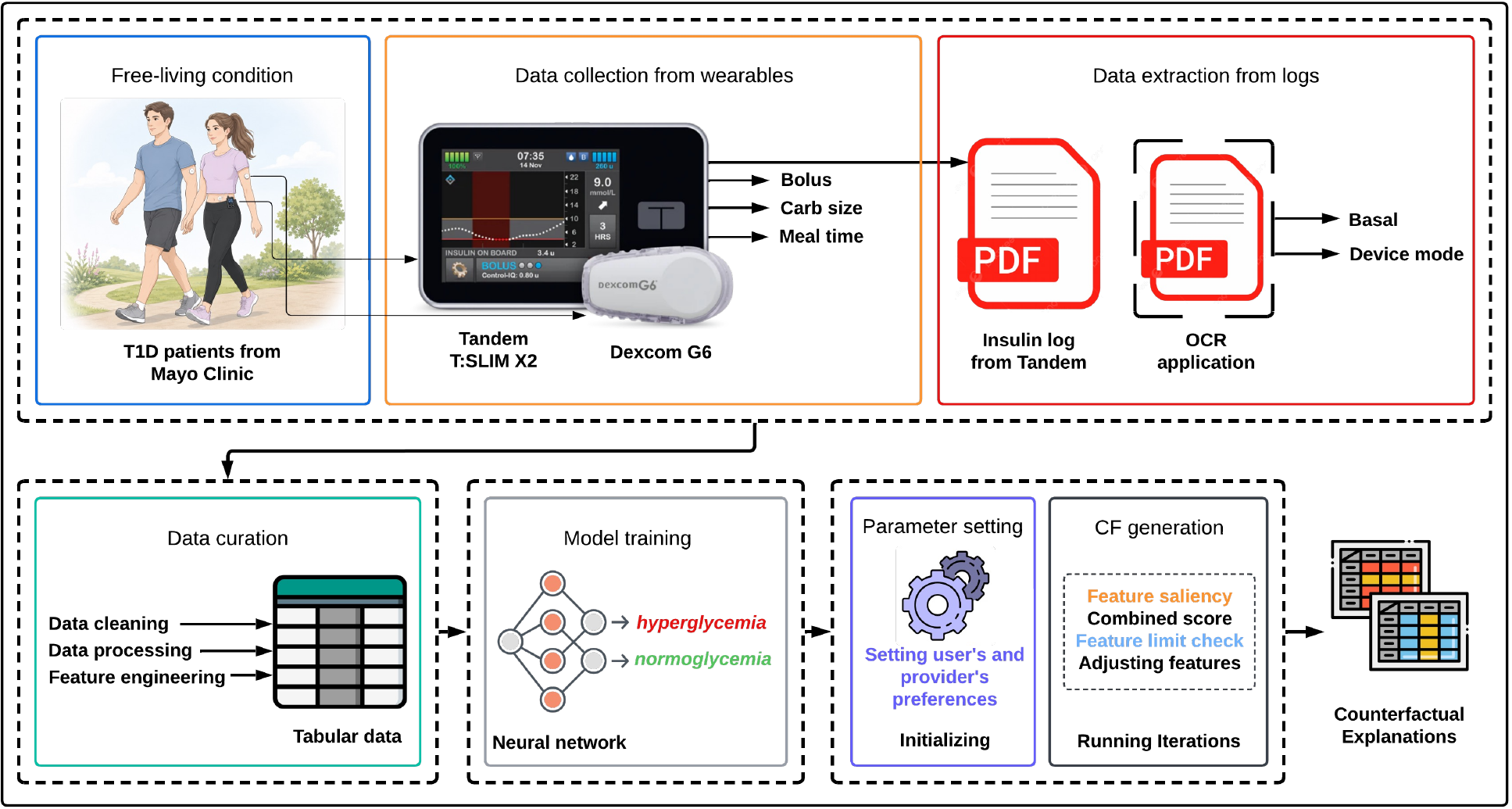}
\caption{GlyTwin framework consists of four phases: data acquisition from CGM sensor and insulin logs, model training for glycemic outcome prediction, counterfactual generation for actionable recommendations, and integration into a dynamic, personalized management pipeline.}
\label{fig:glytwin_dia}
\end{figure*}

\figref{glytwin_dia} illustrates the high level overview of the pipeline used to develop and test the hypothesis behind GlyTwin. Within the pipeline, we define distinct phases for data collection, data curation, model training, and CF generation.

The AZT1D dataset \cite{Khamesian2025AZT1DAR} is constructed by obtaining data from individuals with T1D at the Mayo Clinic Arizona who are on Automated Insulin Delivery (AID) systems. Information such as carbohydrate sizes, bolus doses, basal amounts, and device modes (e.g., sleep or exercise mode) is extracted from the obtained data. Next, we process the data to ensure quality and to derive additional variables. This step includes feature engineering, handling missing values, removing outliers, and formatting data for further analysis.

During the model development phase, a neural network is trained to classify outcomes as hyperglycemic or normoglycemic based on input features. Nonetheless, GlyTwin is model-agnostic and can work with any model regardless of its type.

The CF generation phase begins by initializing GlyTwin with user and provider preference weights ($w_u$ and $w_p$). During each iteration, GlyTwin computes feature saliency to identify the feature that, when perturbed, gives maximum leverage towards normoglycemia. The saliency, combined with the user preference weights, determines the feature to be perturbed first. A perturbation ($\delta$) is then applied to the selected feature and the process—feature selection and adjustment—repeats until a predefined confidence threshold ($\gamma$) towards normoglycemia is achieved. Therefore, the interventions generated by GlyTwin ensure minimal changes aligned with user preferences and lead to the desired outcome.

\section{Data}
\subsection{Data Collection}
Data is collected from $50$ individuals with T1D who visited the Endocrinology Department of Mayo Clinic, Phoenix, AZ between December 2023 and April 2024 as part of their regular treatment (IRB \#23-003065). To ensure consistent insulin delivery algorithm across all subjects, only individuals using Tandem T:SLIM X2 Pump and Dexcom G6 Pro CGM systems were included in the study. For each subject, the data contains approximately $26$ days of recordings collected in free-living settings and includes glucose readings from CGM device, insulin logs, meal carbohydrate sizes, and device modes (regular/sleep/exercise) from the Tandem insulin pump. Next, data from the $50$ individuals (Age: $55.9 \pm 16.5$ years, $28$ female, A1C level: $5.0-8.2\%$, $47$ White and $3$ Hispanic) is processed further for developing and testing GlyTwin. After removing the missing data ($15.8\pm11.72$ hours per individual), the final dataset includes approximately 30,450 hours of glucose readings, basal rates, carbohydrate intakes, and bolus intakes. Table~\ref{azt1} summarizes the demographics of the subjects in AZT1D dataset.

%IRB \#23-003065

\begin{table}[h]
\footnotesize
\centering
\caption{Demographic information of the AZT1D dataset collected from the Mayo Clinic.}
\label{azt1}
{\renewcommand{\arraystretch}{1.4}

\begin{tabular}{>{\raggedright\arraybackslash} C{0.08cm} C{1.2cm} C{0.5cm} C{1.3cm} C{1.3cm} C{1.8cm}}
\toprule
\textbf{\textit{n}} & \textbf{Age (mean$\pm$SD)} & \textbf{Gender (F/M)} & \textbf{A1C (mean$\pm$SD)} & \textbf{YfD (mean$\pm$SD)} & \textbf{Ethnicity (White/Hispanic)} \\
\myrowcolour%
\hline
50 &  55.9$\pm$16.5& 28/22 & 6.73$\pm$0.74 & 31.41$\pm$15.76 & 47/3 \\
\hline
\hline
\end{tabular}}
\end{table}

\subsection{Data Preprocessing}
\begin{figure*}[!t]
    \centering
    % First row
    \subfigure[Basal rate extraction]{
        \includegraphics[width=\textwidth]{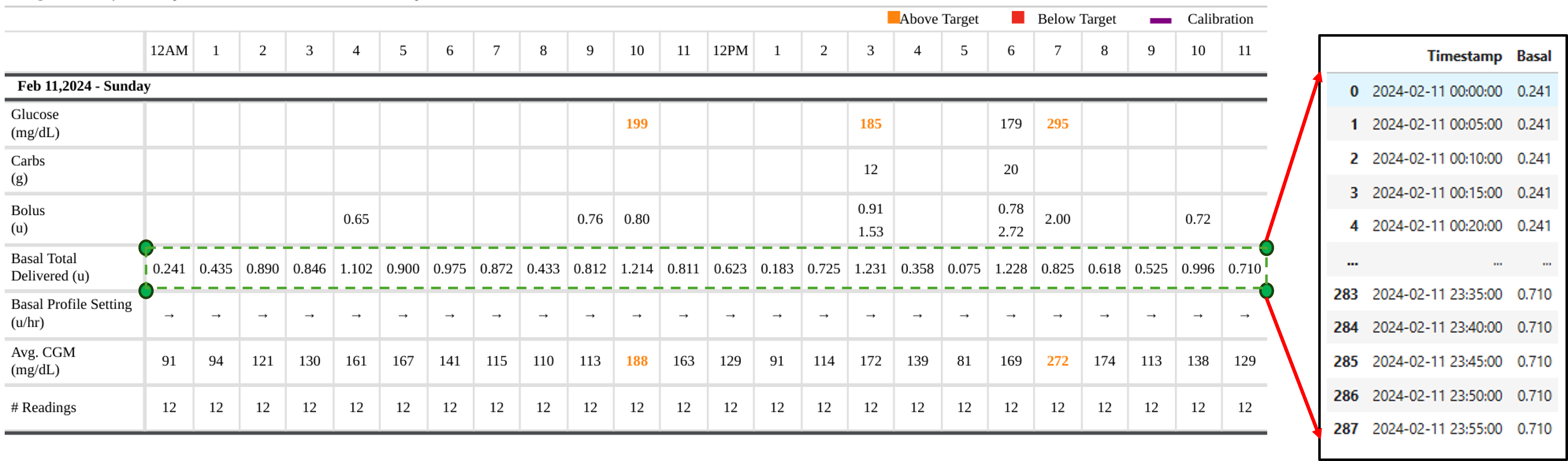}
        \label{fig:basal}
    }
    % Second row
    \subfigure[AID device mode extraction]{
        \includegraphics[width=\textwidth]{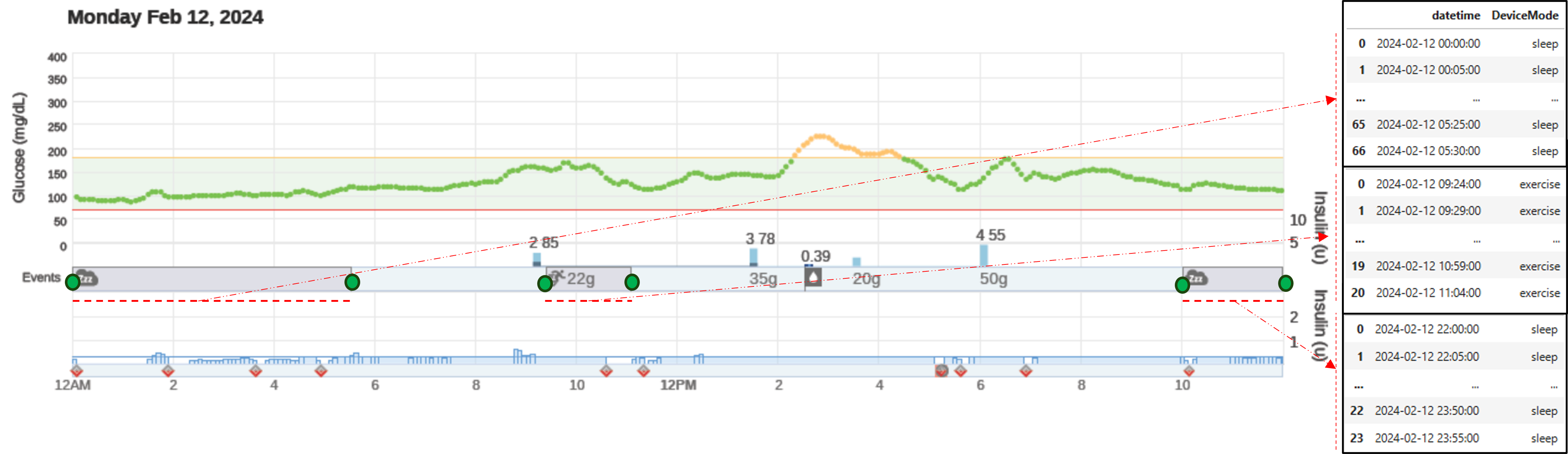}
        \label{fig:mode}
    }
    \caption{Extracting the basal rates and the device modes from the PDFs using OCR and coordinate system.}
    \label{fig:basal_modes}
\end{figure*}

The raw data went through the following preprocessing steps:

\subsubsection{Basal rates and device modes} The hourly basal rates and device modes in the PDF files downloaded from Tandem are extracted by cropping the informative areas and then using an \underline{\textbf{O}}ptical \underline{\textbf{C}}haracter \underline{\textbf{R}}ecognition (OCR) technique. \figref{basal} and \figref{mode} shows the extraction process of basal rates and device modes using OCR and coordinate system.

\subsubsection{Time between meal and food bolus, $\Delta t$} Prior research \cite{Cameron2009ProbabilisticEM,Corbett2019AMH,Peters2017POSTPRANDIALDO} says nearly $32\%$ individuals with T1D bolus after or during the meal which is one of the key reasons behind post-meal hyperglycemia. So, making suggestions on improving $\Delta t$ may play a key role in improving glycemic control. 

To estimate $\Delta t$, first, the timestamps ($t_{fb}$) for food boluses have been identified from the timeseries data. Following two hours of $t_{fb}$, the maximum post-meal glycemic response and its timestamp ($t_{max}$) have been captured. Therefore, $BGL_{max} = \max (BGL[t_{fb}:t_{fb}+120])$ and $t_{max} = \arg\max (BGL[t_{fb}:t_{fb}+120])$. Since, peak time for glucose level after meal is $72\pm23$ minutes, our assumption is that meal timestamp $t_{meal}=t_{max}-72$ \cite{Daenen2010PeaktimeDO}. Hence, we calculate $\Delta t$ using $t_{meal}$ and $t_{fb}$. 

\subsubsection{Total bolus} Total bolus is the sum of all bolus intakes taken between $\min(t_{meal},t_{fb})$ and $t_{max}$.

\subsubsection{Total basal} Sum of all basal units taken between $t_{meal}-90$ and $t_{meal}$ falls under Total basal feature.

\subsubsection{Pre-meal glucose level and slope} The CGM reading at $t_{meal}$ is the pre-meal blood glucose level. A linear trend-line is fitted using the prior 30 minutes' ($t_{meal}-30:t_{meal}$) glucose readings and the pre-meal glucose level slope (or blood glucose change rate every 5 minute) is calculated from it. 

\subsubsection{Filtering out carb sizes}
Oftentimes, individuals with T1D intend to compensate for their high blood sugar levels with additional doses of food boluses instead of administering correction boluses. This scenario leads to some secondary carb sizes in close temporal proximity of the primary one. On those occasions, we take the carb size with maximum value into account and neglect the rest that fall within $\min(t_{meal},t_{fb})$ and $t_{max}$.

\subsubsection{Outlier removal}
Outliers are removed using an Interquartile Range (IQR)–based filter, where values falling outside the range 
\([Q_1 - 3 \times \mathrm{IQR},\; Q_3 + 3 \times \mathrm{IQR}]\) for any feature are excluded \cite{Tukey1977ExploratoryDA}. This removes extreme deviations but retains the majority of the data distribution.

\begin{figure}[!h]
\centerline{\includegraphics[width=\columnwidth]{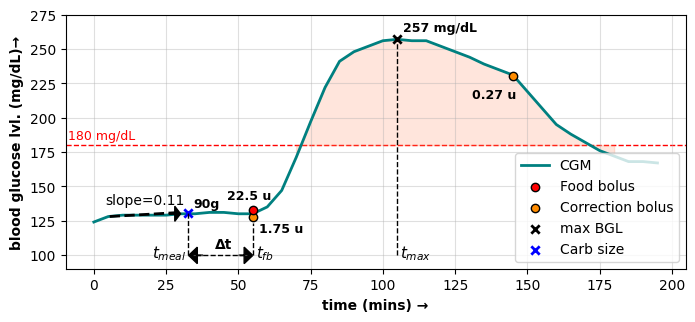}}
\caption{Derivation of different features from the data stream.}
\label{fig:fig2}
\end{figure}

All the above mentioned features and calculations are illustrated in \figref{fig2}. The data processing pipeline leaves us with \textbf{2672} (1220 hyperglycemic, 1452 normoglycemic) factual samples. Simple randomization assigned data to training and testing cohorts using an $85/15$ split at both the subject level $(43/7)$ and datapoint level $(2328/344)$. Two samples are shown in Table~\ref{sample} as examples. Of the eleven features, we consider \textit{Carb size}, \textit{Total bolus}, \textit{$\Delta t$}, and \textit{Pre-meal BGL} as modifiable factors for behavioral modification.

\begin{table*}[!h]
\caption{Examples of processed samples from the dataset.}
\label{sample}
\centering
\scriptsize
{\renewcommand{\arraystretch}{1.1}
\arrayrulecolor{black} % <--- makes all table lines blue

\begin{tabular}{c|c|c|c|c|c|c|c|c|c|c|c}
\toprule
\textbf{Age} & \textbf{Gender} & \textbf{Ethnicity} & \textbf{A1C} & \textbf{Carb size} & \textbf{Total bolus} & \textbf{$\Delta t$}  & \textbf{Mode} & \textbf{Total basal} & \textbf{Pre-meal BGL slope} & \textbf{Pre-meal BGL} & \textbf{Outcome} \\
\myrowcolour%
\hline 
61 & F & White & 6.7 & 20 & 7.57 & -5 & regular & 2.475 & 2.943 & 129 & \textcolor{cyan}{normoglycemia} \\ 
\hline 
32 & F & Hispanic & 5 & 35 & 5.83 & 15 & regular & 0.357 & 1.457 & 134 & \textcolor{red}{hyperglycemia} \\ 
\bottomrule
\end{tabular}}
\end{table*}

\section{Experimental setup}

\subsection{Classifier Details}
The fully-connected (MLP) binary classifier for hyperglycemia classification is described in Table~\ref{tab:classifier}. 

Classifier hyperparameters are tuned using a systematic grid search. Specifically, multiple architectural and training hyperparameters are explored to identify the configuration that maximizes performance across validation loss, accuracy, F1-score, and AUC. The grid search varied:

\begin{itemize}
    \item \textbf{Layer sizes:} [64, 32, 16], [128, 64, 32], [64, 32, 32], [32, 32, 16], and [64, 64, 32]
    \item \textbf{Dropout rates:} [0.2, 0.2, 0.2], [0.2, 0.15, 0.15], [0.2, 0.15, 0.1], [0.15, 0.15, 0.1] and [0.15, 0.1, 0.1]
    \item \textbf{Learning rates (Adam optimizer):} 1e-2, 1e-3 and 1e-4
    \item \textbf{Batch sizes:} 4, 8 and 16
\end{itemize}

Each combination was trained for $130$ epochs, with validation performance on a held-out test set ($15$\% data on both subject level and data point level) recorded at the end of training. The configuration yielding the highest validation AUC was selected as the optimal model. Results are generated using the optimal model.

All experiments are performed using a single compute node with access to 12 CPU cores, 24 GiB of RAM, and a single NVIDIA L40 GPU for hardware acceleration.

\begin{table}[h]
\scriptsize
\centering
\caption{Classifier specifications for hyperglycemia prediction.}
\label{tab:classifier}
{\renewcommand{\arraystretch}{1.4}

\begin{tabular}{>{\raggedright\arraybackslash}p{2cm} p{5.7cm}}
\toprule
\textbf{Layer} & \textbf{Description} \\
\myrowcolour%
\hline
\textbf{Input Layer} & Dense, $64$ neurons, \textit{leakyrelu} (0.1) activation, \textit{HeNormal} initializer, Batch normalization, Dropout rate: $0.2$ \\
\hline
\textbf{Hidden Layer 1} & Dense, $32$ neurons, \textit{leakyrelu} (0.1) activation, \textit{HeNormal} initializer, Batch normalization, Dropout rate: $0.15$ \\
\myrowcolour%
\hline
\textbf{Hidden Layer 2} & Dense, $16$ neurons, \textit{leakyrelu} (0.1) activation, \textit{HeNormal} initializer, Batch normalization, Dropout rate: $0.1$ \\
\hline
\textbf{Output Layer} & Dense, $1$ neuron, \textit{sigmoid} activation \\
\myrowcolour%
\hline
\textbf{Optimizer} & Adam, learning rate: $1e-4$ \\
\hline
\textbf{Dataset Split} & $85/15$ train/test split -- both at subject and sample level \\
\myrowcolour%
\hline
\textbf{Training} & $130$ epochs, batch size: $4$ \\
\hline
\textbf{Performance} & Accuracy: $82.63\%\pm0.007\%$, F1-score: $0.784\pm0.009$, AUC: $0.837\pm0.01$ $[$over 4 trials$]$\\
\hline
\hline
\end{tabular}}
\end{table}
\subsection{Parameter set}
Algorithm~\ref{alg} contains multiple parameters that need to be initialized prior to running it. For preventing hyperglycemia, we set target class $y'=$\textit{ normoglycemia} and the corresponding confidence ($\gamma$) at $0.6$ as this value of $\gamma$ maintains a delicate balance between validity and proximity. We set the maximum iterations to $N=400$, and consider four features modifiable: \textbf{Carb size}, \textbf{Total bolus}, \textbf{$\Delta t$}, and \textbf{Pre-meal BGL}. Their corresponding perturbation size, $\delta$ values are $5$ grams, $0.5$ unit, $5$ minutes, and $10$ mg/dL, respectively. We personalize the minimum and maximum values for the modifiable features according to individual subject, but set the minimum and maximum pre-meal blood glucose levels to $100$ and $170$ mg/dL, respectively. When we compare GlyTwin against other techniques, unless mentioned otherwise, both the physician's preference ($w_p$) and user's preference weights ($w_u$) are set to $1$ for all modifiable features.

\subsection{Baselines} 
We have identified the following techniques to compare against GlyTwin.

\subsubsection{DiCE}DiCE \cite{mothilal2020dice} identifies a set of CFs by optimizing for proximity, diversity and sparsity. 

\subsubsection{Optbinning}In Optbinning \cite{Navas-Palencia2021Counterfactual}, CFs are generated by optimizing binning rules to modify input features with an aim to find the shortest path to a target class.

\subsubsection{CFNOW}CFNOW \cite{DEOLIVEIRA2023} searches an optimal point close to the factual point where the classification differs from the original. CFNOW performs greedy optimization for metrics like speed, coverage, distance, and sparsity.
 
\subsubsection{NICE}CFs by NICE \cite{Brughmans2021NICEAA} are not necessarily adversarial data points but nearby instances in the data manifold that reflects the desired outcome.

\subsection{Validation Metrics}
Validating the CFs has been a persistent challenge \cite{Miller2017ExplanationIA}. We assess the CFs using standard metrics found in the literature:

\textbf{\textit{Validity}} assesses whether the produced CFs genuinely belong to the desired class \cite{Guo2021CounterNetET}. High validity indicates the technique’s effectiveness in generating valid CF examples. Like \cite{Barbiero2021EntropybasedLE}, a simulation-aided method is designed to estimate the validity of the CFs.
\begin{equation*}
    \textit{validity} = \frac{\#|f(X_T^*) \neq f(X_T)|}{\|CF\|}
\end{equation*}

The simulator used to estimate validity is an XGBoost model trained with real data. With a max-depth of $2$, learning rate of $0.62$, $12$ estimators and $85\%$ training data, the XGBoost simulator achieves $80.23\%$ accuracy and $0.798$ F1-score.

\textbf{\textit{LoopInsighT1 validity}} GlyTwin generated CFs are further validated using an off-the-shelf glucose level simulator called LoopInsighT1 \cite{Peuscher2024AMO} which is developed using UVA Padova physiological model of insulin pharmacodynamics, carbohydrate absorption, and glucose regulation \cite{Man2014TheUT}. This way, a CF is considered valid if the resulting simulated glycemic response stays within-target glycemic range.

\textbf{\textit{Nearest Neighbor Test (NN Test)}} validates the effectiveness of the CFs by comparing them against historical data to determine their likely outcomes (e.g., hyperglycemia or normoglycemia) based on past similar instances. We implement it using a k-nearest neighbor (k-NN) algorithm where $k=7$ and the accuracy is $83.8\%$. This is very similar to the yNN test in \cite{Laugel2019TheDO} and \cite{Williams2024RethinkingDM}

\textbf{\textit{Proximity}} is the $L_2$ norm distance between $X_T$ and ${X_T}^*$. A low \textit{Proximity} ensures we are making small change to the factual sample by preserving the details and not over-correcting the user \cite{Wachter2017CounterfactualEW}.
\begin{equation}
\textit{proximity} = \sqrt{\sum_{i=1}^{m_{cont.}} \left( \frac{x_T^{*i}}{\|x_T^{*i}\|_2} - \frac{x_T^{i}}{\|x_T^{i}\|_2} \right)^2}
\end{equation}
where:
$m_{cont.}$ refers to the number of continuous features.

\textbf{\textit{Sparsity}} is the average number of feature changes per CF. A low sparsity ensures better user understanding of the CFs \cite{Guo2021CounterNetET}.
\begin{equation}
\textit{sparsity} = \frac{\sum_{X_T^*\in CF}^{}\sum_{i=1}^{d} \mathbbm{1}(x_T^{*i} \neq x_T^i)}{\|CF\|} 
\end{equation}

\textbf{\textit{Violations}} quantifies how frequently non-modifiable features (e.g. age, gender, insulin etc.) are changed. A good CF technique will have fewer violations per CF and promote fairness \cite{Chen2024CounterfactualFT}. If $d_{\text{\st{mod}}}$ is the number of non-modifiable features-
\begin{equation}
    \textit{violations} = \frac{\sum_{X_T^*\in CF}^{}\sum_{k=1}^{d_{\text{\st{mod}}}} \mathbbm{1}(x_T^{*k} \neq x_T^k)}{\|CF\|}
\end{equation}

\textbf{\textit{Plausibility}} estimates the fraction of explanations that  fall within the feature ranges derived from the data \cite{Guidotti2022}-
\begin{equation*}
    \textit{plausibility}=\frac{\sum_{X_T^*\in CF}^{}\mathbbm{1}(\text{dist}(X_T^*)\subseteq\text{dist}(X))}{\|CF\|}
\end{equation*}
where, dist($X_T^*$) and dist($X$) represent the distribution of feature values in the CF instances $X_T^*$ and in the training data, respectively. $\|CF\|$ is the total number of CF instances.

Finally, average \textbf{\textit{feature diversity}} has been calculated using the following formula-
\begin{equation*}
    \text{Average diversity for feature }k = \frac{\sum_i\sum_j|{x_i}^k-{x_j}^k|}{\|CF\|}, i\neq j
\end{equation*}

\begin{table*}[!b]
\small
\caption{Evaluating the CFs on AZT1 Data: GlyTwin outperforms others in validity, NN test, proximity (except CFNOW, which has low validity and NICE, which identifies CFs from the training data), violations, plausibility and achieves comparable results in sparsity. Results are reported as Mean$\pm$SD over 4 trials. Arrows indicate if higher ($\uparrow$) or lower ($\downarrow$) values are better.}
\label{evaluation}
\centering
{\renewcommand{\arraystretch}{1.1}
\begin{tabular}{p{1in}|C{0.62in}C{0.66in}C{0.62in}C{0.62in}C{0.62in}C{0.62in}C{0.67in}}
\toprule
 \multirow{2}{*}{Method} &\multicolumn{7}{c}{\textbf{Validation metrics}} \\
& \cellcolor{blue!18}validity $\uparrow$ & \cellcolor{orange!18}LoopInsighT1 validity $\uparrow$ & \cellcolor{gray!18}NN test $\uparrow$ & \cellcolor{red!25}proximity $\downarrow$ & \cellcolor{green!25}sparsity $\downarrow$ & \cellcolor{yellow!25}violations $\downarrow$ & \cellcolor{teal!25}plausibility $\uparrow$ \\
\midrule 
GlyTwin bi-$\Delta t$ & \textbf{0.858$\pm$0.004} & \textbf{0.9} & \textbf{0.873$\pm$0.004} & 0.199$\pm$0.007 & 2.566$\pm$0.019 & \textbf{0.0$\pm$0.0} &  \textbf{1.0$\pm$0.0}  \\
GlyTwin one-$\Delta t$ & \textbf{0.858$\pm$0.019} & 0.869 & 0.835$\pm$0.004 & 0.302$\pm$0.021 & 2.274$\pm$0.087 & \textbf{0.0$\pm$0.0} &  \textbf{1.0$\pm$0.0}  \\
DiCE \cite{mothilal2020dice} & 0.852$\pm$0.017 & 0.862 & 0.831$\pm$0.028 & 0.243$\pm$0.008 & 1.667$\pm$0.057 & \textbf{0.0$\pm$0.0} &  0.999$\pm$0.167  \\
Optbinning \cite{Navas-Palencia2021Counterfactual} & 0.2$\pm$0.0 & 0.438 & 0.169$\pm$0.0 & 0.309$\pm$0.0 & 3.062$\pm$0.0 & \textbf{0.0$\pm$0.0} & \textbf{1.0$\pm$0.0} \\
CFNOW \cite{DEOLIVEIRA2023} & 0.702$\pm$0.046 & 0.769 & 0.557$\pm$0.034 & \textbf{0.078$\pm$0.0} & 2.229$\pm$0.174  & 1.062$\pm$0.138 &  0.771$\pm$4.261  \\
NICE \cite{Brughmans2021NICEAA} & 0.785$\pm$0.0 & 0.838 & 0.838$\pm$0.0 & 0.111$\pm$0.0 & \textbf{1.654$\pm$0.0} & 0.377$\pm$0.0 &  0.908$\pm$0.0  \\
\bottomrule
\end{tabular}}
\end{table*}
\begin{table*}[t!]
\caption{Examples interventions made on hyperglycemic pre-meal contexts using GlyTwin.}
\label{explanations}
\centering
\footnotesize
{\renewcommand{\arraystretch}{1.1}
\begin{tabular}{p{4.8in}|p{1.6in}}
\toprule
\textbf{Pre-meal context}& \textbf{Intervention}  \\
\myrowcolour%
\hline 
 Tracey is a 69 year old white individual with T1D whose A1C is 7.2. She entered 20g carb size in her Tandem insulin pump and then got a bolus intake of 2.79 unit. Over the last 90 minutes, she took 0.95 unit basal. Somehow, she ate 45 minutes later when her pre-meal blood glucose reading was 149 mg/dL and blood glucose change rate was 0.91 mg/dL every 5 minute and pump was set at 'regular' mode. Eventually, she experienced post-meal hyperglycemia.
 & She could have prevented it just by reducing her carb intake by 10g and waiting until pre-meal glucose drops to 129 mg/dL. \\ 

\hline 
 Rachel is 67 with an A1C of 6.6. Her last bolus shot was 4.1 units, taken before her meal which had 41 g carb size. She took 0.46 units of basal insulin over the last 90 minutes, and her pre-meal blood sugar level became 113 mg/dL. She experienced hyperglycemia after the meal. & With 7.69 units of bolus and 100 mg/dL pre-meal blood glucose level, Rachel could have prevented hyperglycemia.   \\ 

\bottomrule
\end{tabular}}
\end{table*}
Table \ref{evaluation} summarizes the  CF interventions generated by 
\section{Results}
\subsection{Quality of the counterfactuals} Now, we evaluate the quality of the CF interventions generated by different baseline methods and compare them against our two configurations: GlyTwin bi–$\Delta t$ and GlyTwin one–$\Delta t$. Table \ref{evaluation} outlines that both GlyTwin variants achieve the highest validity scores among all methods evaluated. GlyTwin bi–$\Delta t$ attains a validity of $0.858$, matching the one-$\Delta t$ model and outperforming DiCE, NICE, CFNOW and Optbinning by margins of approximately $0.6\%$, $7.3\%$, $15.6\%$, and $65.8\%$ respectively. Note that, no technique gets a perfect score because the explanations are evaluated by an external simulator and not by the corresponding classifier. GlyTwin bi–$\Delta t$ also achieves the best LoopInsightT1 validity, with a score of $0.9$, indicating that the CFs adhere well to domain-relevant temporal constraints.

According to the NN test, which validates CFs against historical data, GlyTwin bi–$\Delta t$ again outperforms all other techniques with a score of $0.873$, maintaining a margin by more than $4\%$ over DiCE and at least $3.5\%$ above NICE, the strongest baseline. The one–$\Delta t$ variant also performs strongly with $0.835$.

In terms of proximity, GlyTwin bi–$\Delta t$ achieves the closest CFs to the factual samples with an average distance of $0.199$, surpassing DiCE, Optbinning and NICE. Although CFNOW achieves a lower proximity score ($0.078$), this comes at the cost of poor validity and increased violations, reflecting its tendency to modify only a narrow subset of features, primarily A1C, as CFNOW considers all features modifiable.

Across all methods, violations remain lowest for both GlyTwin versions, which achieve a perfect score of 0.0, tying with DiCE and Optbinning. Similarly, both GlyTwin variants attain the highest possible score ($1.0$), indicating that their generated CFs remain entirely within the data manifold.

Finally, regarding sparsity, GlyTwin requires modifying more features on average (bi–$\Delta t$: $2.566$, one–$\Delta t$: $2.274$) relative to DiCE, NICE, and particularly CFNOW, which achieves the sparsest CFs ($2.229$). As noted previously, CFNOW’s apparent advantage in sparsity stems from exclusively altering A1C, which negatively affects its validity and plausibility.

Table~\ref{explanations} presents two interventions with brief storytelling and how they may appear in a clinical setting.

\begin{figure*}[!h]
    \centering
    
        \parbox{\textwidth}{%
            \centering
            % First row
            \subfigure[Classification by subject age]{
                \includegraphics[width=0.48\textwidth]{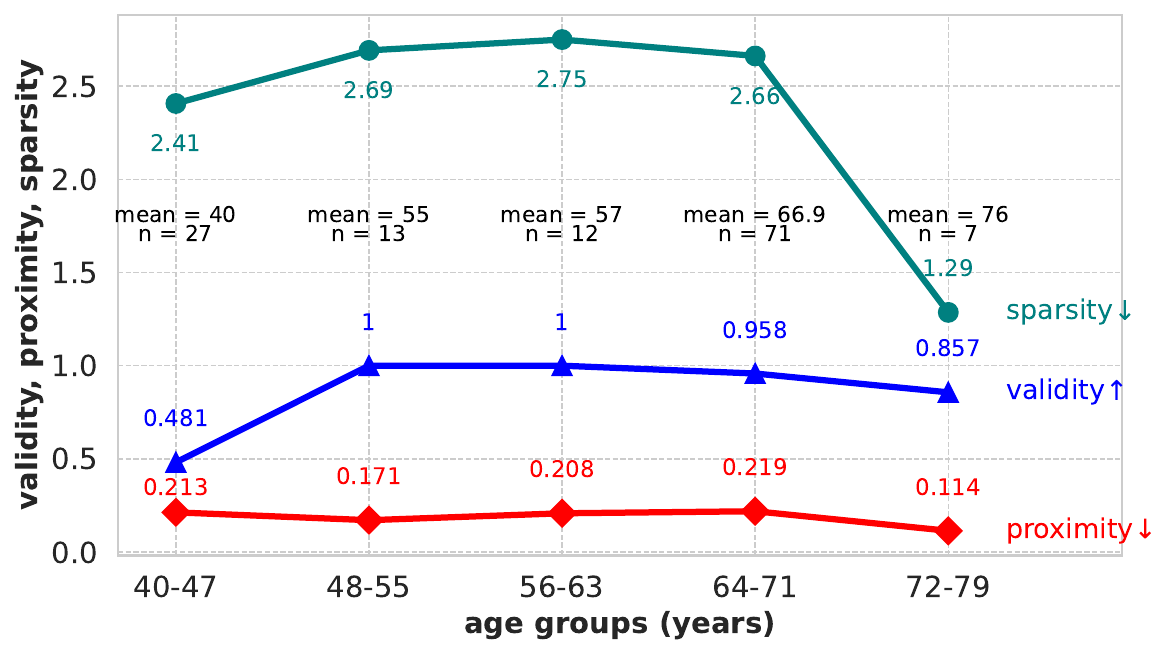}
                \label{age}
            }
            \hfill
            \subfigure[Classification by subject gender]{
                \includegraphics[width=0.48\textwidth]{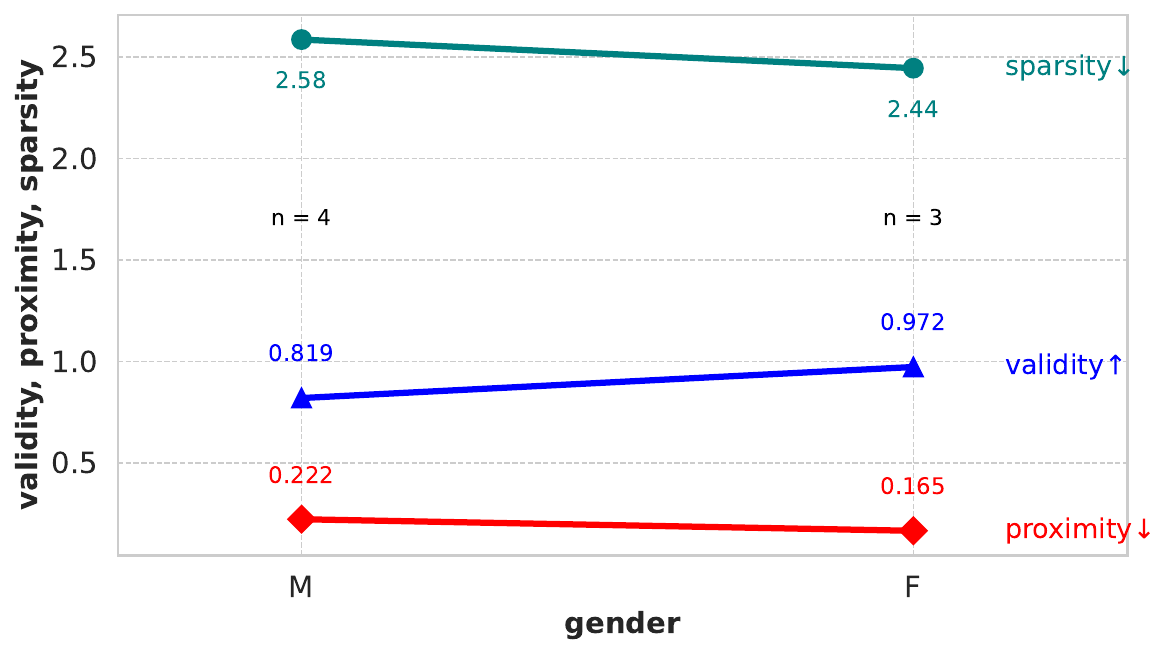}
                \label{gender}
            }

            \vspace{0.8em}

            % Second row
            \subfigure[Classification by subject A1C]{
                \includegraphics[width=0.48\textwidth]{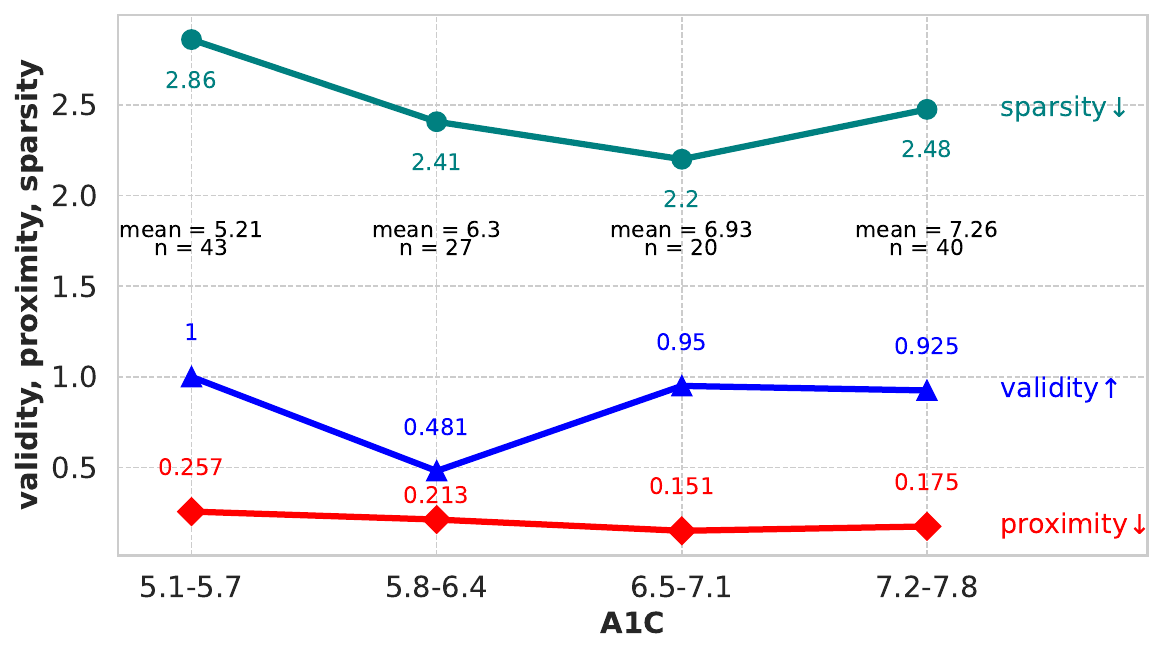}
                \label{A1C}
            }
            \hfill
            \subfigure[Classification by years from diagnosis]{
                \includegraphics[width=0.48\textwidth]{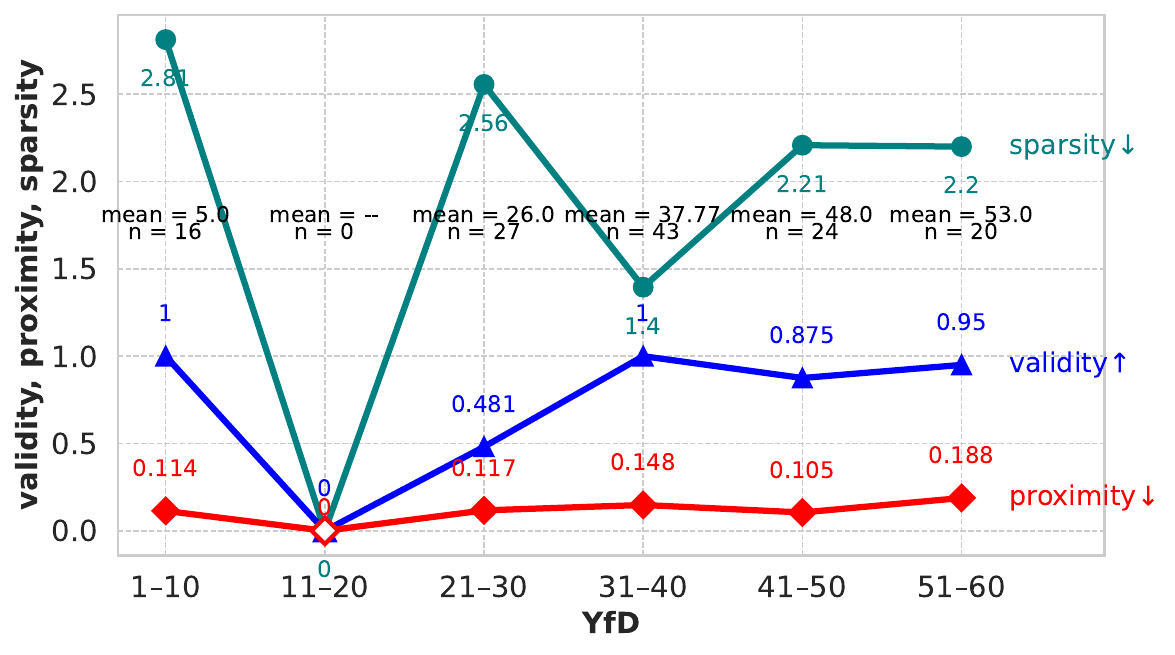}
                \label{yfd}
            }
        }%
    
    \caption{Categorizing the performance of GlyTwin based on subject age, gender, A1C and years from diagnosis (YfD).}
    \label{subfigures}
\end{figure*}

\subsection{Classifying the results}
To further understand how GlyTwin behaves across different subgroups, we classified the evaluation metrics by age, gender, A1C, and years from diagnosis (YfD), as shown in Fig.~\ref{age}–\ref{yfd}. Our analyses reveal consistent patterns in model performance as well as subgroup-specific behavior. 

GlyTwin maintains strong and stable performance across all age groups, with validity values ranging from $0.481$ to $1.0$ (Fig.~\ref{age}). Younger individuals (ages $48-55$ and $56-63$) achieve the highest validity ($1.0$), suggesting that their postprandial glucose excursions are more predictable and the simulator validates the CF interventions more reliably. In contrast, the $40-47$ group shows lower validity ($0.481$), despite a relatively large sample size ($n=27$), indicating greater physiological variability that may hinder precise simulation responses. The oldest group ($72-79$) attains validity of $0.857$, suggesting improved reliability compared to the youngest cohort, although the sample size ($n=7$) may partially influence this result.

Proximity remains nearly constant across age ranges, varying only between $0.114$ and $0.219$, reflecting GlyTwin’s ability to generate changes that remain close to the original data regardless of age. Sparsity is also stable, with values clustered between $2.29$ and $2.75$. Notably, sparsity slightly decreases for older adults ($72-79$, sparsity = $1.29$), which suggests that simpler CFs are sufficient for this subgroup.

As shown in Fig.~\ref{gender}, GlyTwin performs strongly across genders, with males achieving a validity of $0.819$ and females achieving $0.972$. This difference suggests that the model generalizes slightly better for female subjects under the simulator’s evaluation. Conversely, proximity is lower for females ($0.165$) than for males ($0.222$), implying that CFs for female subjects remain more tightly aligned with the original test instances. Sparsity remains similar across genders, modifying a comparable number of features for both subgroups.

Results classified by HbA1c values (Fig.~\ref{A1C}) reveal a clear and intuitive pattern: subjects with lower A1C ($5.1-5.7\%$) exhibit higher validity ($1.0$) than those with higher A1C levels ($7.2-7.8\%$, validity = $0.925$). This suggests that individuals with tighter glycemic control may have fewer actionable directions for improvement, which makes valid CF generation slightly challenging. Proximity steadily decreases as A1C increases, from $0.257$ in the lowest group to $0.175$ in the highest, meaning that the CFs for individuals with poorer glycemic control deviate less from the original glucose trajectory. Sparsity shows modest variation, peaking at $2.86$ for the lowest A1C group and decreasing to $2.48-2.2$ in intermediate groups, reflecting fewer required modifications to achieve valid CFs.

Classification by YfD (Fig.~\ref{yfd}) also reveals meaningful subgroup differences. Individuals within the $1-10$ year range achieve the highest validity ($1.0$), suggesting that their GlyTwin is more effective for those who are in their early years in T1D. Subgroup $11-20$ has no sample. Sparsity also fluctuates across YfD categories, peaking at $2.88$ for the $1-10$ year group and decreasing in subsequent ranges, indicating that individuals with longer disease duration may require fewer behavioral adjustments to reach the desired glycemic outcome. Proximity remains consistently low across all YfD groups, demonstrating GlyTwin’s stability in generating CFs that remain close to the original observations regardless of disease duration.

\begin{figure}[!b]

\centerline{\includegraphics[width=0.82\columnwidth]{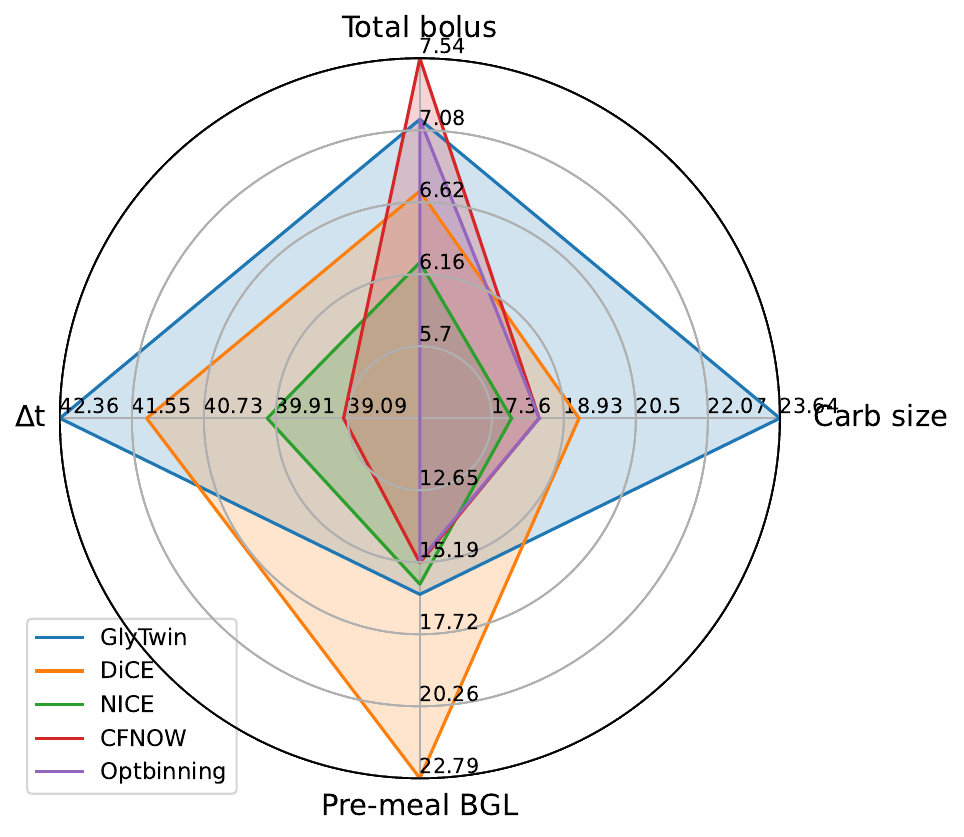}}
\caption{Comparison of feature diversity among CFs produced using different techniques.}
\label{fig:diversity}
\end{figure}

\subsection{Diversity of the counterfactuals}
Next, we assess GlyTwin's performance in terms of feature diversity. Although GlyTwin does not optimize for improving feature diversity, as shown in the radar plot of \figref{diversity}, it exhibits better feature diversity for two out of the four modifiable features, with the two exception being \textit{Total bolus} and \textit{$\Delta t$}. Higher feature diversity means GlyTwin is exploring the entire data while generating the CF interventions and not limited to a specific subset of values for each feature.

\subsection{Alignment with the preference weights}

\begin{figure}[!h]
\centerline{\includegraphics[width=\columnwidth]{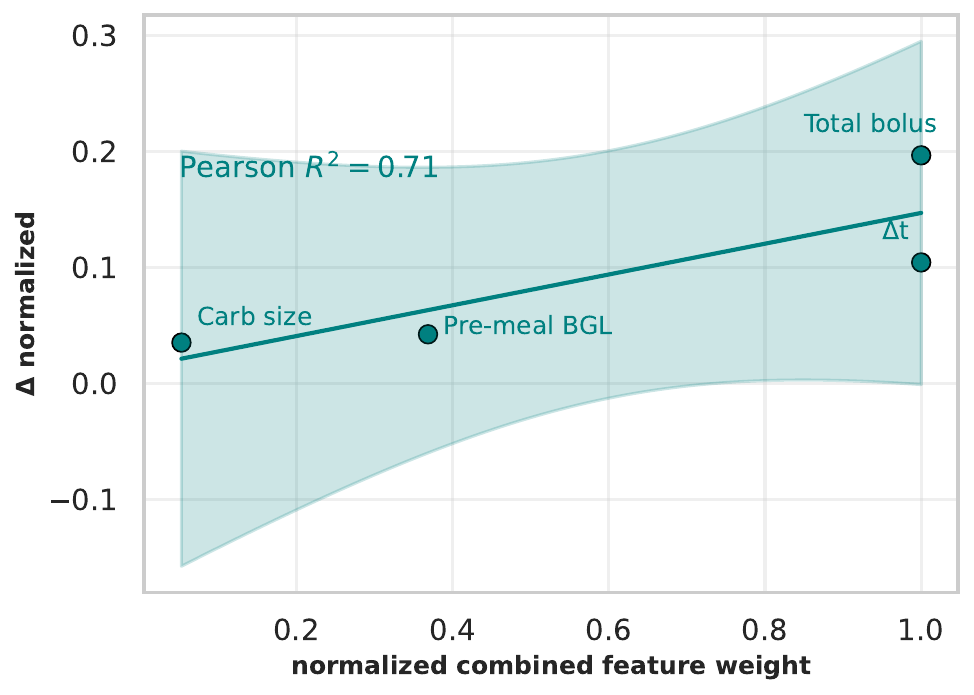}}
\caption{Preference alignment analysis of GlyTwin.}
\label{fig:pre_alignment}
\end{figure}

To understand if CFs from GlyTwin align with stakeholders' preferences, we set physician's preference weights ($w_p$) to [0, 0.9, 0.9, 0] and randomly assign user's preference weights ($w_u$) to [0.1, 1, 1, 0.7], respectively, for carb size, total bolus, $\Delta t$ and pre-meal blood glucose level. We monitor the absolute changes in the corresponding features, normalize both the combined preference weights and the feature changes and plot them in \figref{pre_alignment}. The two variables are positively correlated ($r^2 = 0.71$), despite the impact of feature saliency.

\begin{figure*}[!h]
    \centering
    
    \subfigure[\label{fig:v1}]{{\includegraphics[scale=0.37]{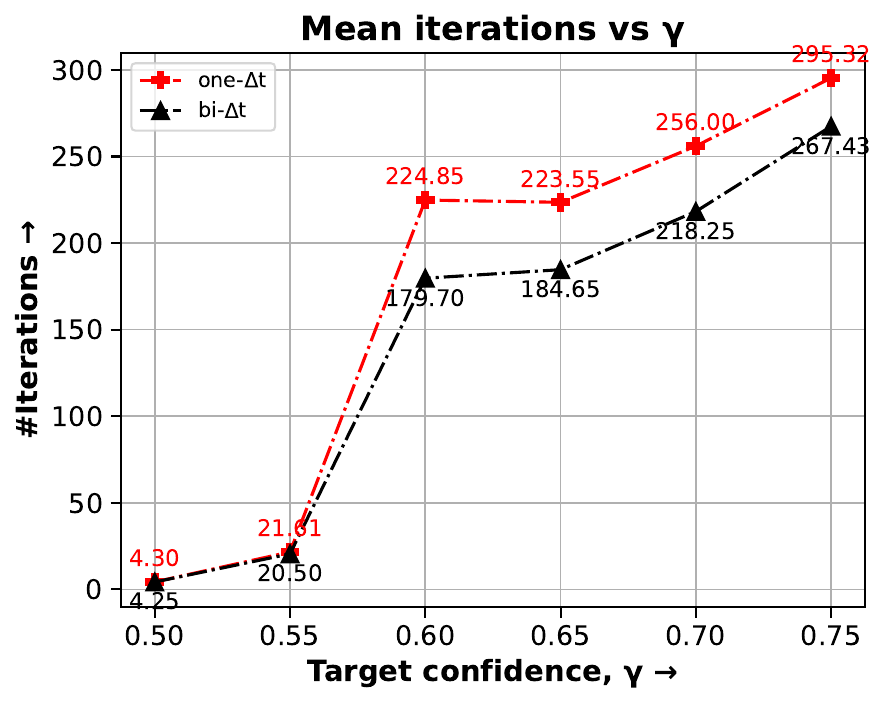} }}%
    \quad
    \subfigure[\label{fig:v2}]{{\includegraphics[scale=0.37]{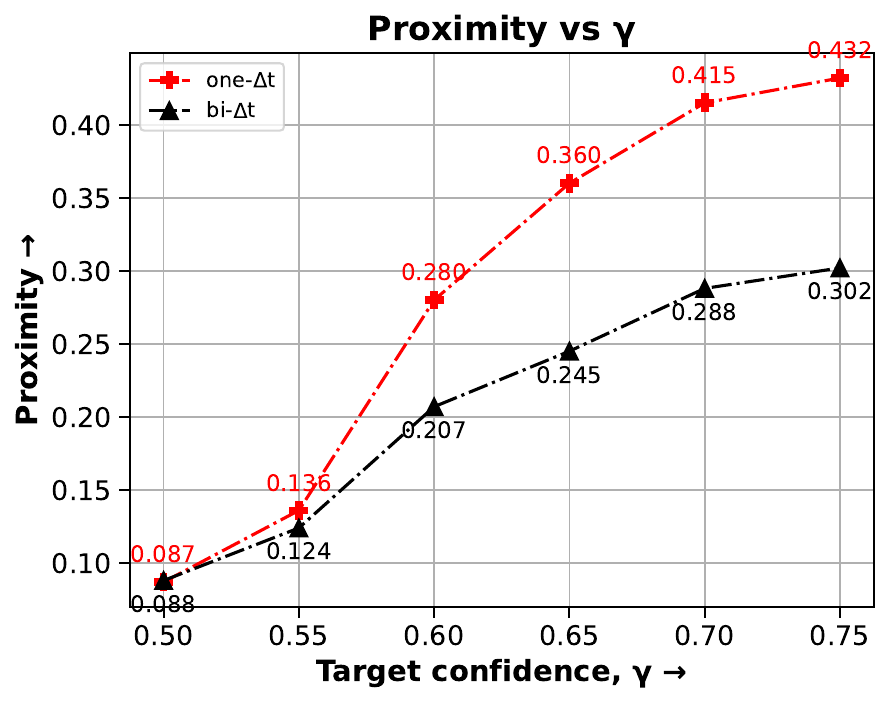} }}%
    \quad
    \subfigure[\label{fig:v3}]{{\includegraphics[scale=0.37]{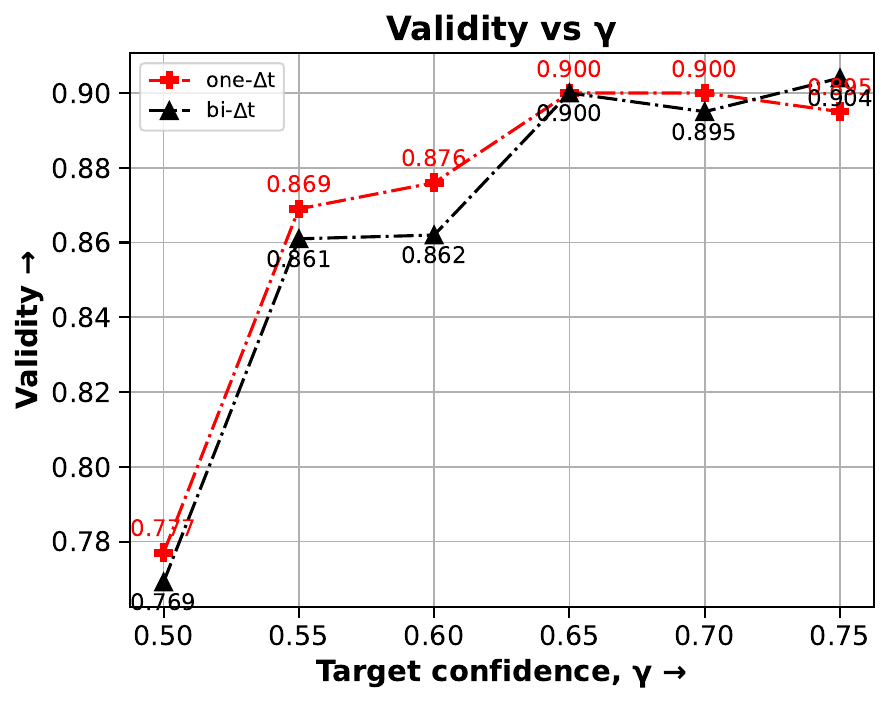} }}%
    \caption{Results of ablations studies performed using GlyTwin. The ablation studies include understanding (a) how the number of required iterations, (b) proximity, and (c) validity changes as we set different target confidence for achieving normoglycemia.}
    \label{abl1}%
\end{figure*}

\section{Discussion}

In this section, we will demonstrate some experiments conducted on the GlyTwin algorithm. Basically, we want to answer- what happens when certain parameters of GlyTwin are modified?

\subsection{Impact of target probability $\gamma$}
When a higher target probability $\gamma$ is set for normoglycemia, it takes longer for GlyTwin to reach the target. Therefore, GlyTwin has to operate for additional iterations to converge. At the same time, converging to a higher target probability requires making more changes to the original factual sample which results in a higher proximity score. Furthermore, these additional iterations and proximity scores solidifies the CF's position in the simulated validation with a higher validity score. \figref{v1}, \figref{v2}, and \figref{v3} depict how number of iterations, proximity score and validity, respectively, change as we increase $\gamma$ from $0.50$ to $0.75$. For $\gamma=0.75$ validity of the generated CFs reaches as high as $0.904$ as determined by the external simulator.

\subsection{Impact of perturbation size $\delta$} In this segment, we vary perturbation size $\delta$ from $5\%$ to $25\%$ of the feature ranges and run GlyTwin algorithm to understand the impact of $\delta$ on the validation metrics. The summary of the impact is depicted in Table~\ref{delta_scores}.

Validity peaks when $\delta \text{ is } 25\%$ of the feature range. Other than that it remains close to $0.8$. With higher $\delta$, the generated CFs are placed further from the factual samples. Thus, there is a trade-off between validity and proximity. While higher $\delta$ improves validity, it simultaneously worsens proximity slightly ranging from $0.152$ at $\delta= 5\%$ to $0.184 \text{ at } \delta = 25\%$ of feature range. Sparsity is somewhat robust to perturbation size as it remains stable and close to $1$ feature per CF across all $\delta$ values with narrowly increasing from $1.038$ to $1.211$. Quite understandably, runtime decreases significantly as $\delta$ increases, dropping from $2$ seconds at $\delta = 5\%$ to $0.6$ second at $\delta = 20\%$, before increasing slightly at $\delta = 25\%$.

Therefore, $\delta$ needs to be large enough to make a difference and converge faster but not so large that it introduces incorrect changes or takes the CFs far from the factual samples.

\begin{table}[t]
\centering
\footnotesize
\renewcommand{\arraystretch}{1}
\caption{Scores for Different Delta Values}
\begin{tabular}{|c|c|c|c|c|}
\toprule
\textbf{$\delta$ (\%)} & \textbf{Validity} & \textbf{Proximity} & \textbf{Sparsity} & \textbf{Runtime (s)} \\ 
\midrule \midrule
5 & 0.746 & 0.075 & 1.038 & 2.07$\pm$2.58 \\ 

10 & 0.762 & 0.08 & 1.038 & 1.43$\pm$3.12 \\ 

15 & 0.793 & 0.087 & 1.023 & 1.01$\pm$2.57 \\ 

20 & 0.823 & 0.092 & 1.015 & 0.57$\pm$0.28 \\ 

25 & 0.837 & 0.114 & 1.211 & 0.83$\pm$1.19 \\ 
\bottomrule
\end{tabular}
\label{delta_scores}
\end{table}

\subsection{How long does GlyTwin take?}

\begin{figure}[!h]
\includegraphics[width=\columnwidth]{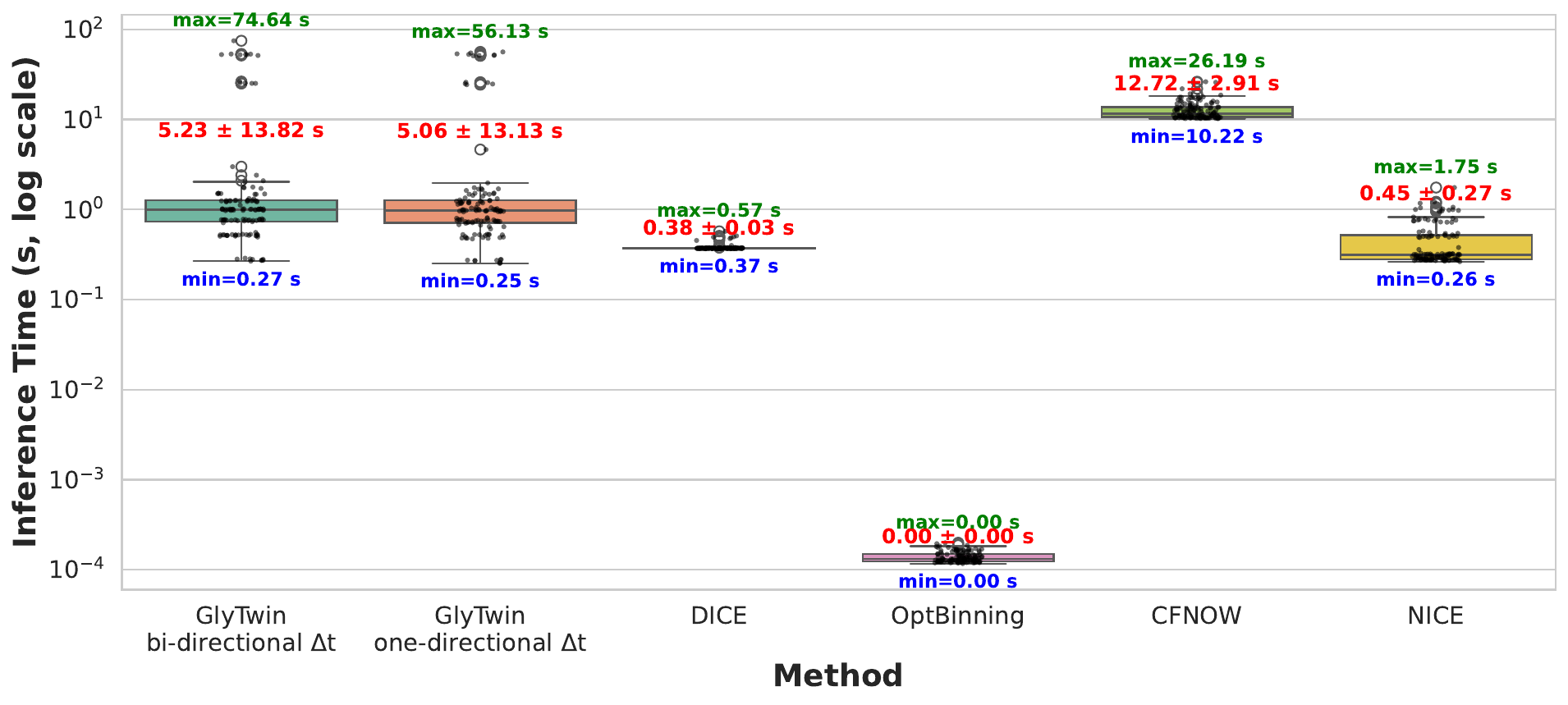}
\caption{Runtime comparison between different methods. GlyTwin takes roughly $5$ seconds on average to produce a behavioral intervention to prevent hyperglycemia.}
\label{fig:time}
\end{figure}

Since some features are already at the terminal values or deemed non-modifiable, not all factual samples can converge to corresponding CF samples, regardless of the number of iterations. However, GlyTwin achieves $100\%$ convergence and takes approximately $5.2$ seconds on average to generate a CF. \figref{time} compares the time required to generate a CF across different methods.

\subsection{Limitations of GlyTwin}
Although GlyTwin achieves promising results in many performance metrics, we have identified three limitations associated with the current state of the GlyTwin technology.

\subsubsection{Clinical validation of the interventions}
The main limitation of GlyTwin is that the generated CFs have not been clinically tested. While the XGBoost-based emulator provides a convenient way to approximate how feature adjustments may influence predicted outcomes, its validation power is inherently limited. Although the intervention inputs and simulator were trained on data drawn from similar behavioral and physiological distributions, this alone does not guarantee true physiological fidelity—particularly when CFs involve behavioral adjustments that extend beyond frequently observed patterns. As a predictive model, XGBoost captures statistical associations within the data manifold but hardly encodes mechanistic glucose–insulin dynamics, and therefore should be interpreted as offering predictive consistency rather than physiologically grounded validation.

To strengthen the reliability of our validation, we additionally evaluated the GlyTwin-generated CF interventions using LoopInsighT1 open-source simulator for type 1 diabetes that incorporates physiological glucose–insulin dynamics, insulin absorption models, and carbohydrate appearance curves. This complementary validation provides a more mechanistic perspective on whether the CF recommendations yield plausible postprandial trajectories. While this does not replace the need for formal clinical evaluation, it offers an additional layer of confidence that the generated interventions are clinically plausible.

However, neither of these analyses takes into account user's adherence to the provided behavioral intervention. Therefore, the true impact of the interventions in preventing post-meal hyperglycemia also relies on the individuals' ability to follow them. Designing an AI-driven intervention is challenging, but transitioning it to real settings is even more challenging with issues like scalability, application development, regulatory compliance, user adaptability, and maintenance. Therefore, this study focuses solely on the development phase and simulation-aided validations of GlyTwin.

\subsubsection{Small data size}
Although data were collected from 100 individuals, only 50 could be included in this study due to substantial missing CGM segments, lack of recorded carbohydrate sizes for individuals using primarily quick boluses, and the need for manual extraction of basal rates and device modes from inconsistently formatted pump PDFs. These constraints limited the usable sample size in the present analysis. Our ongoing efforts to improve the data extraction pipeline will help add more subjects in future work to enhance generalizability.

\subsubsection{Estimation of $\Delta t$}
We acknowledge the limitations associated with GlyTwin’s peak-based estimation of meal timing following \cite{Daenen2010PeaktimeDO}. Although postprandial peak has been used in several prior studies for retrospective meal identification in CGM data \cite{Ahmed2023MealtimeDB, Camerlingo2023138LBAR}, this approach is sensitive to sensor noise, behavioral variability, and physiological delays. In individuals using AID systems, glucose trajectories may also be shaped by automated correction boluses or micro-adjustments in basal insulin, introducing further variability in peak timing and increasing the uncertainty of this heuristic. Nonetheless, because GlyTwin is designed for deployment in real-world, free-living conditions—where meal announcements are often inconsistent, or unreliable—robust operation under timestamp uncertainty is a practical requirement. We emphasize that future work will have more precise evaluation once datasets with ground-truth meal timing, like the T1DEXI \cite{Riddell2023ExaminingTA}, are incorporated for further analysis.

\subsubsection{Suggested $\Delta t$ values}
Another limitation is that the CFs generated by GlyTwin often suggest delaying the bolus intake, i.e., taking the bolus after meals. While we do not have a concrete explanation for such interventions, our assumption is that delaying the bolus intake could help avoid insulin stacking. Insulin stacking occurs when multiple insulin doses are taken in close proximity and increases the risk of hypoglycemia. Insulin stacking is highly prevalent in our data, and delayed-bolus recommendations are strongly correlated with factual instances where insulin stacking occurred.

In our analysis, GlyTwin suggests delaying insulin by 10 minutes for $9$ hyperglycemic instances $(7\%)$ and by 5 minutes for $14$ instances $(10.8\%)$. For $78$ hyperglycemic instances $(60\%)$, GlyTwin instead recommends administering the bolus earlier, whereas for the remaining $29$ instances $(22.3\%)$, no change in bolus timing is suggested. Notably, recommendations to delay the meal bolus most frequently arise in cases where insulin stacking occurred in the original data. In such situations, GlyTwin favors administering a single, comparatively larger bolus dose with at most a 10-minute delay to prevent further insulin stacking.

GlyTwin algorithm is very flexible. Therefore, in the \textit{one-directional $\Delta t$} version, the $\Delta t$ (bolus–meal timing) feature is constrained such that modifications are only permitted when the direction of change indicates an earlier bolus administration relative to the original instance.

\subsubsection{Higher computation cost} Since GlyTwin is an iterative algorithm, it has higher computation overhead. It is also a multi-step technique consisting of model training and adjusting the perturbations. Therefore, it is not fast compared to methods that train both the classifier and the generative model jointly and invokes the pipeline during inference \cite{Guo2021CounterNetET}. Time complexity of the GlyTwin algorithm for creating one CF is $O(N*|I_{mod}|)$, where, $N$ refers to the maximum number of allowed iterations and $|I_{mod}|$ is the number of modifiable features.

We are exploring integration with low-latency models and optimization of the saliency computation step and model's inference time. Our preliminary profiling indicates that targeted acceleration of the feature perturbation loop and model-querying process can yield substantial gains in runtime efficiency. We also plan to introduce adaptive step sizes to reduce the number of required iterations, caching gradients and reusing saliency scores across iterations to minimize redundant computations without compromising explanation quality.

\subsubsection{Device-specific generalizability} A limitation of this study is that all participants used the Tandem t:slim X2 insulin pump paired with the Dexcom G6 Pro CGM. This single-system inclusion criterion maintained consistent insulin-delivery algorithms and uniform data extraction across individuals, however, it also restricts the generalizability of GlyTwin to other pump–sensor combinations (e.g., Medtronic with Guardian, Omnipod with Dexcom) that differ in their delivery logic and data structure. As a result, GlyTwin’s performance should be interpreted within the context of this specific device ecosystem. Future work will include data from a broader range of pump and CGM systems to evaluate cross-device applicability.

\subsubsection{Multimodal analysis} Although digital twin systems are inherently multimodal, GlyTwin is intentionally designed around a minimal set of highly actionable features like carbohydrate intake, insulin dosing, timing, and pre-meal glucose, which are reliably available in free-living settings. This design choice reflects constraints of free-living settings, where additional modalities like physical activity and sleep data are often inconsistently recorded or unavailable. Future work will include multimodal inputs like activity and sleep, to enable more comprehensive intervention recommendations.

GlyTwin does not integrate physical activity, which is known to influence insulin sensitivity and postprandial glucose dynamics; as a result, the current framework focuses solely on food- and insulin-related factors and may not fully generalize to contexts where activity plays a substantial metabolic role. GlyTwin also does not take insulin-on-board (IOB), maximum bolus caps, or minimum inter-bolus intervals into consideration. We have plans add data from wearable activity sensors, IOB modeling to as part of multimodal analysis.

Finally, the present study serves as a proof-of-concept demonstration of GlyTwin rather than a full validation across diverse datasets. Several simplifying assumptions, like peak-based meal-time estimation and reliance on resource-heavy iterative computations, narrow the scope of GlyTwin. Hence, our findings should be interpreted within this limited context, with the understanding that additional refinements may be needed as GlyTwin is tested in broader, benchmark datasets.

\section{Future Validation Plans} To support eventual clinical deployment, we outline a structured, multi-phase validation plan to progressively evaluate the safety, usability and effectiveness of GlyTwin’s interventions.
\subsubsection{Phase 1 — Provider-facing evaluation}
A clinician-facing interface including the GlyTwin algorithm will be provided to endocrinologists and nurse practitioners to generate CF recommendations on pre-recorded data. This phase will assess GlyTwin’s ability to reduce predicted hyperglycemia risk, correctness, workflow integration and perceived clinical utility.
\subsubsection{Phase 2 — Supervised pilot study}
Contingent on Phase 1 outcomes, a small-scale pilot study will be conducted in which individuals with type 1 diabetes follow GlyTwin-generated interventions under endocrinologist supervision, while their postprandial glucose curve will be closely monitored. Presence of experts will ensure that insulin- and carbohydrate-related recommendations remain safe and clinically appropriate.
\subsubsection{Phase 3 — Controlled clinical trial}
GlyTwin will subsequently be deployed within a mobile application for a larger controlled study comprising a baseline observation period followed by a GlyTwin-assisted intervention period. Outcomes will include time-in-range (TIR), frequency of hyperglycemic and hypoglycemic events, postprandial excursion metrics, and participant-reported measures of usability and satisfaction.

Together, these phases provide a stepwise path from in silico evaluation to supervised clinical testing and to GlyTwin’s safety and effectiveness assessment, consistent with established diabetes management standards and research protocols.

\section{Conclusion}
GlyTwin refines the traditional digital twin technology with CFs and empowers stakeholders to participate in the CF generation process. GlyTwin is built and tested on real data collected in uncontrolled settings. Our analysis demonstrates that the generated CFs are valid, fair, realistic, and minimal, eventually matching or outperforming existing techniques in producing quality CFs. Our next venture includes overcoming the aforementioned limitations and have it trialed with subjects with T1D in a clinical setting following the designed outlines.

\section*{Data Availability}
The dataset used in this study is publicly available at: \href{https://data.mendeley.com/datasets/gk9m674wcx/1}{\textcolor{blue}{https://data.mendeley.com/datasets/gk9m674wcx/1}}

\section*{Acknowledgment}
This work was supported in part by the National Institute of Diabetes and Digestive and Kidney Diseases of the National Institutes of Health under grant T32DK137525, the National Science Foundation under grant IIS-2402650, and the Mayo Clinic and Arizona State University Alliance for Health Care Collaborative Research Seed Grant Program. Any opinions, findings, conclusions, or recommendations expressed in this material are those of the authors and do not necessarily reflect the views of the funding organization.

\section*{References}
\bibliographystyle{IEEEtran}

\end{document}

\title{Supplementary Materials for GlyTwin}

\maketitle

\begin{figure*}[b!]
    \centering
    \begin{subfigure}{0.35\textwidth}
        \centering
        \includegraphics[width=\linewidth]{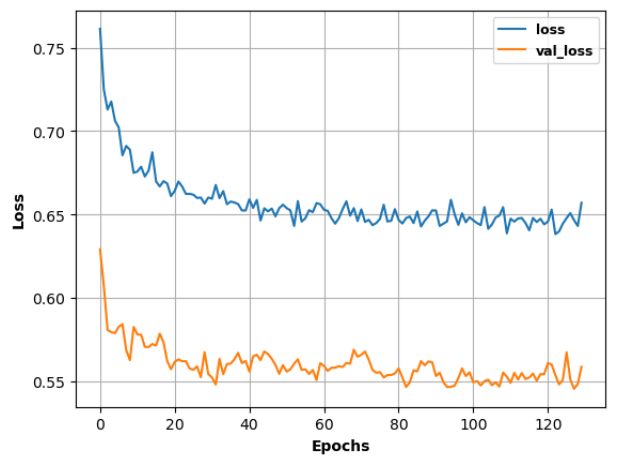}
        \caption{}
        \label{tc1}
    \end{subfigure}
    \quad
    \begin{subfigure}{0.35\textwidth}
        \centering
        \includegraphics[width=\linewidth]{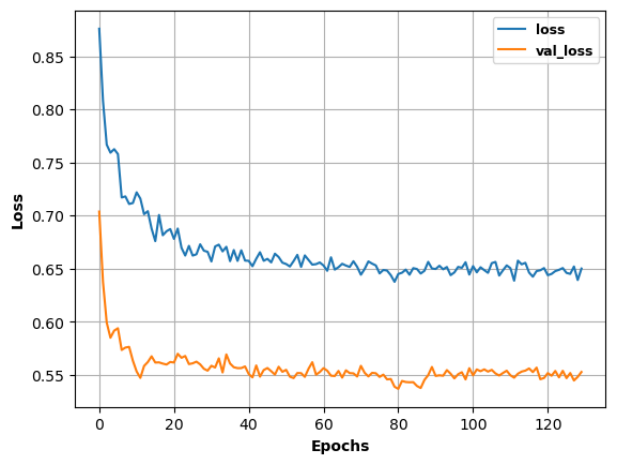}
        \caption{}
        \label{tc2}
    \end{subfigure}
    \quad
    \begin{subfigure}{0.35\textwidth}
        \centering
        \includegraphics[width=\linewidth]{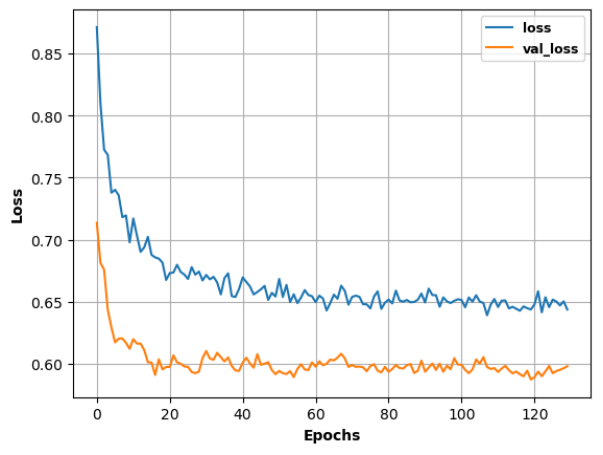}
        \caption{}
        \label{tc3}
    \end{subfigure}
    \quad
    \begin{subfigure}{0.35\textwidth}
        \centering
        \includegraphics[width=\linewidth]{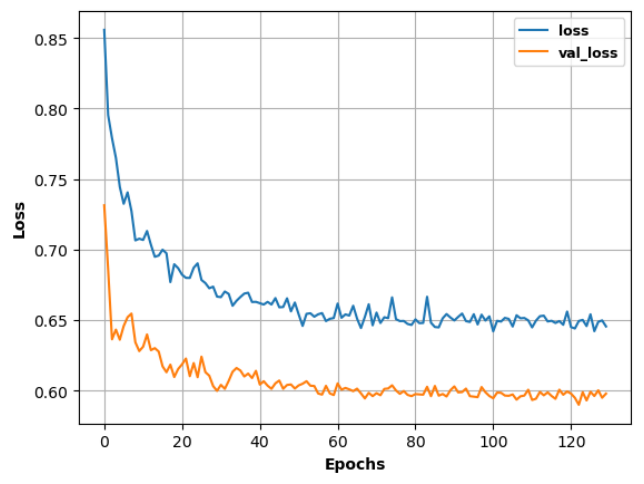}
        \caption{}
        \label{tc4}
    \end{subfigure}

    \caption{Training curves for different seed values.}
    \label{fig:tc_all}
\end{figure*}

\section{Training curves, Calibration and Per-class performance}
\subsection{MLP Classifier}
Figures~\ref{tc1}-\ref{tc4} shows the training curves for the MLP classifier.

Table~\ref{tab:cls_results} shows the per-class performance of the MLP classifier.

\begin{figure}[!h]
\centerline{\includegraphics[width=0.85\columnwidth]{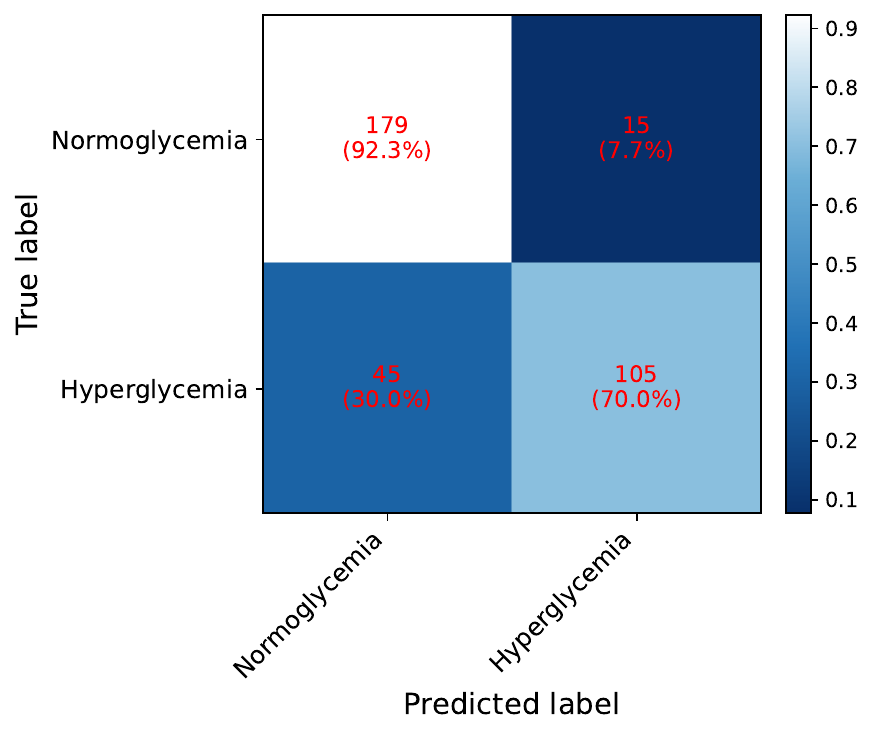}}
\caption{Confusion matrix for the classifier.}
\label{conf_mat}
\end{figure}

Figure~\ref{conf_mat} is the confusion matrix for the classifier.

\textbf{\textit{Class imbalance:}} Class imbalance was addressed by applying inverse-frequency class weighting during model training, which ensured that minority-class samples contributed proportionally more to the loss function. Specifically, class weights were computed using the balanced scheme in \texttt{sklearn}, where each class weight is inversely proportional to its prevalence in the training data, and these weights were added into the training objective to mitigate bias toward the majority class.

\begin{table}[h]
\centering
\caption{Per-class classification results for the classifier}
\label{tab:cls_results}
\begin{tabular}{lcccc}
\toprule
Class & Precision & Recall & F1-score & Support \\
\midrule
normoglycemia & 0.80 & 0.92 & 0.86 & 194 \\
hyperglycemia & 0.88 & 0.70 & 0.78 & 150 \\
\midrule
Accuracy &  &  & 0.83 & 344 \\
Macro Avg & 0.84 & 0.81 & 0.82 & 344 \\
Weighted Avg & 0.83 & 0.83 & 0.82 & 344 \\
\bottomrule
\end{tabular}
\end{table}

\subsection{XGBoost simulator}
The XGBoost simulator was calibrated using a grid search method in which key hyperparameters were systematically varied within predefined ranges. The grid explored maximum tree depth values in the range [$1, 12$], learning rates spanning [$0.001, 0.99$], and numbers of estimators across [$1, 100$]. The search was optimized using the F1-score. Table~\ref{xgboost} shows class-wise performance for the XGBoost simulation.

\begin{table}[h!]
\centering
\caption{Per-class performance metrics for the XGBoost simulator}
\label{xgboost}
\begin{tabular}{lccc}
\hline
\textbf{Class} & \textbf{Precision} & \textbf{Recall} & \textbf{F1-score} \\
\hline
Class 0 (normoglycemia) & 0.8088 & 0.9072 & 0.8552 \\
Class 1 (hyperglycemia) & 0.8475 & 0.8067 & 0.8266 \\
\hline
\end{tabular}
\end{table}

\subsection{NN test}
This NN test uses \textit{Euclidean distance} (\textit{Minkowski} with $p=2$) to find the $7$ nearest neighbors, gives them \textit{uniform (equal) voting weights}, automatically selects the neighbor search algorithm with a \textit{leaf size of $30$} for efficiency. Table~\ref{tab:nntest_results} shows class-wise performance for the NN-test.

\begin{table}[h]
\centering
\caption{Per-class classification results for NN-test}
\label{tab:nntest_results}
\begin{tabular}{lcccc}
\toprule
Class & Precision & Recall & F1-score & Support \\
\midrule
normoglycemia & 0.83 & 0.88 & 0.85 & 194 \\
hyperglycemia & 0.84 & 0.78 & 0.81 & 150 \\
\midrule
Accuracy &  &  & 0.83 & 344 \\
Macro Avg & 0.84 & 0.83 & 0.83 & 344 \\
Weighted Avg & 0.84 & 0.84 & 0.84 & 344 \\
\bottomrule
\end{tabular}
\end{table}

\section{Baseline Configuration} Table~\ref{tab:cfg-all-methods} contains configurations for baseline methods: DiCE, NICE, CFNOW and Optbinning.

\begin{table*}[b!]
\centering
\caption{Configuration of different baseline methods}
\label{tab:cfg-all-methods}
\resizebox{\textwidth}{!}{%
\begin{tabular}{lcccc}
\toprule
\textbf{Component} &
\textbf{DiCE} &
\textbf{NICE} &
\textbf{CFNOW} &
\textbf{Optioning} \\
\midrule

Total CFs per factual &
\texttt{1} &
-- &
-- &
\texttt{n\_cf = 1} \\

Desired class &
\texttt{opposite} &
-- &
-- &
-- \\

Distance metric &
-- &
\texttt{HEOM} &
-- &
-- \\

Optimization objective &
-- &
\texttt{proximity} &
-- &
-- \\

Justified counterfactuals &
-- &
\texttt{True} &
-- &
-- \\

Time limit &
-- &
-- &
\texttt{10 s} &
-- \\

Number of CFs generated &
-- &
-- &
\texttt{1} &
\texttt{1} \\

Base classifier (model) &
-- &
-- &
-- &
\texttt{LogisticRegression (C = 0.0001)} \\

Model accuracy &
-- &
-- &
-- &
\texttt{80.52\%} \\

Scorecard construction &
-- &
-- &
-- &
\begin{tabular}[c]{@{}l@{}}
\texttt{Scorecard(} \\
\quad \texttt{binning\_process = binning\_process,} \\
\quad \texttt{estimator = estimator,} \\
\quad \texttt{scaling\_method = "min\_max",} \\
\quad \texttt{scaling\_method\_params = \{0, 100\},} \\
\quad \texttt{reverse\_scorecard = True)}
\end{tabular} \\

Features allowed to vary /
Actionable features &
\begin{tabular}[c]{@{}l@{}}
\texttt{CarbSize}, \\
\texttt{TotalBolus}, \\
\texttt{del\_t}, \\
\texttt{PreMeal\_BGL}
\end{tabular} &
-- &
-- &
\begin{tabular}[c]{@{}l@{}}
\texttt{CarbSize}, \\
\texttt{TotalBolus}, \\
\texttt{del\_t}, \\
\texttt{PreMeal\_BGL}
\end{tabular} \\

Permitted range source &
Patient-specific min/max &
-- &
-- &
-- \\

Permitted range (per feature) &
\texttt{[min, max] on matching subset} &
-- &
-- &
-- \\

Weights /
Hyperparameters &
\texttt{proximity\_weight = 0.5} &
Library defaults &
-- &
-- \\

Maximum number of changes &
-- &
-- &
-- &
\texttt{max\_changes = 4} \\

Hard constraints &
-- &
-- &
-- &
\texttt{diversity\_values} \\

Other hyperparameters &
Library defaults &
Library defaults &
-- &
-- \\

\bottomrule
\end{tabular}%
}
\end{table*}

\section{LoopInsighT1 Simulator}
\begin{figure*}[!t]
\centerline{\includegraphics[width=\textwidth]{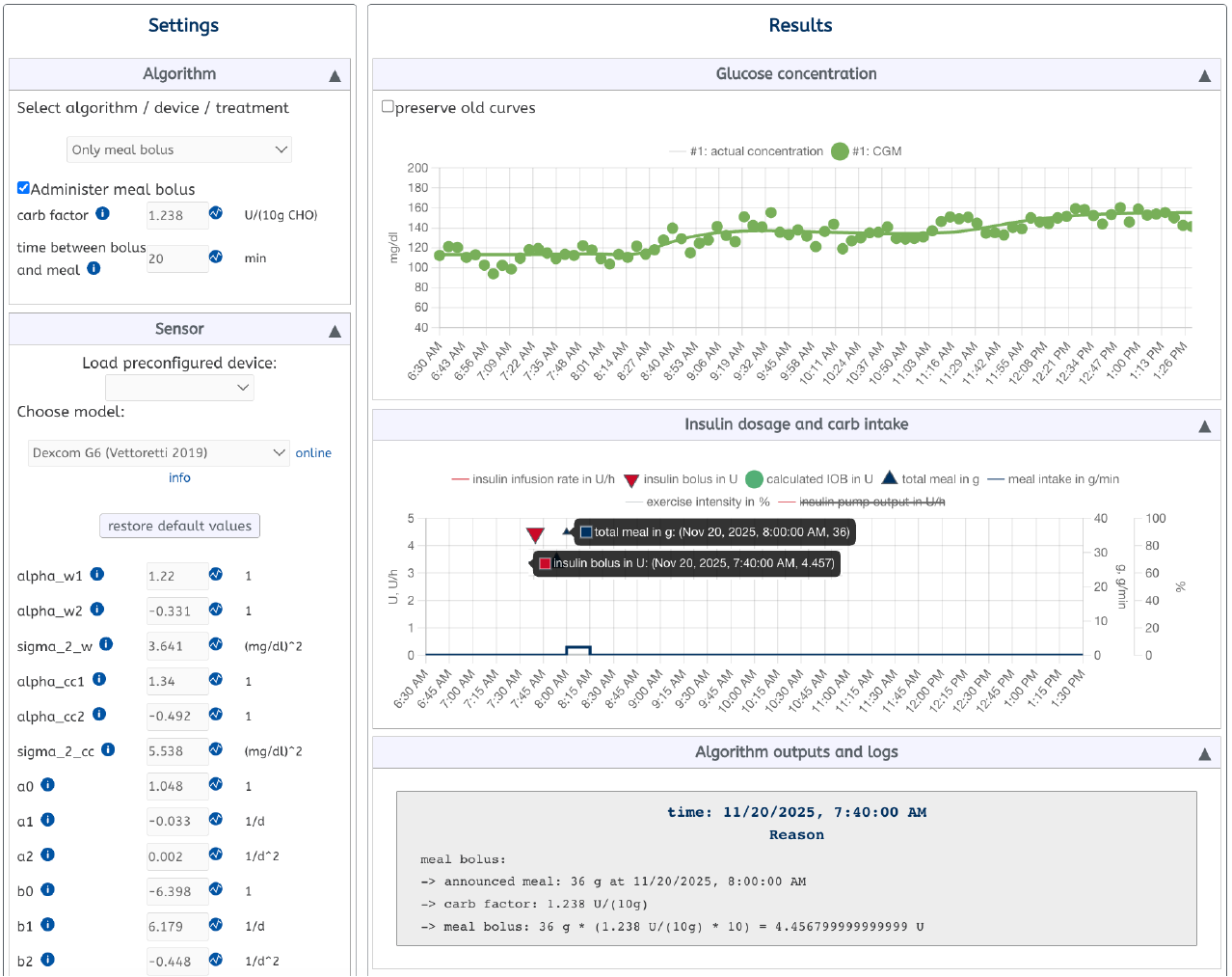}}
\caption{BGL simulation with LoopInsighT1 simulator for a counterfactual intervention.}
\label{loopinsight1}
\end{figure*}

Below is an example of validating an intervention using the simulator. All the details from the interventions (Table~\ref{loopinsighttable}) are fed to the simulator as input and the resulting glycemic response is monitored (Figure~\ref{loopinsight1}).

\begin{table}[h]
\centering
\caption{A counterfactual intervention to be fed to the LoopInsighT1 simulator}
\label{loopinsighttable}
\begin{tabular}{ll}

\hline
\textbf{Feature} & \textbf{Value} \\
\hline
Age & 76 \\
Gender & F \\
Ethnicity & White \\
A1C & 6.8 \\
Carb size & 36 \\
Total bolus & 4.46 \\
$\Delta t$ & -20 \\
Mode & regular \\
Total basal & 0.5485 \\
Pre-meal BGL slope & 2.83 \\
Pre-meal BGL & 113 \\
\hline
\end{tabular}
\end{table}

\section{Video links}
Video description of how the basal rates were extracted using OCR: \href{https://tinyurl.com/4e2t6auv}{\textcolor{blue}{tinyurl.com/4e2t6auv}}

Video description of how the device modes were extracted using a coordinate system: \href{https://tinyurl.com/494jhu9e}{\textcolor{blue}{tinyurl.com/494jhu9e}}